\documentclass[a4paper,fleqn]{cas-dc}

\usepackage[numbers]{natbib}

\usepackage{todonotes}

\usepackage[acronym, nonumberlist]{glossaries}
\makeglossaries
\setacronymstyle{long-short}
\newacronym[longplural=cluster validity indices]{cvi}{CVI}{cluster validity index}
\newacronym{adr}{ADR}{automated demand response}
\newacronym{ai}{AI}{artificial intelligence}
\newacronym{amy}{AMY}{actual meteorological year}
\newacronym{bau}{BAU}{business-as-usual}
\newacronym{beps}{BEPS}{building energy performance simulation}
\newacronym{bes}{BES}{building energy system}
\newacronym{bess}{BESS}{battery energy storage system}
\newacronym{caidi}{CAIDI}{Customer Average Interruption Duration Index}
\newacronym{cdd}{CDD}{cooling degree days}
\newacronym{cop}{COP}{coefficient of performance}
\newacronym{der}{DER}{distributed energy resource}
\newacronym{dhw}{DHW}{domestic hot water}
\newacronym{dod}{DoD}{depth-of-discharge}
\newacronym{dp}{DP}{dynamic programming}
\newacronym{drl}{DRL}{deep reinforcement learning}
\newacronym{dr}{DR}{demand response}
\newacronym{dtw}{DTW}{Dynamic Time Warping}
\newacronym{eh}{EH}{electric heater}
\newacronym{ercot}{ERCOT}{Electric Reliability Council of Texas}
\newacronym{ess}{ESS}{energy storage system}
\newacronym{eui}{EUI}{energy use intensity}
\newacronym{eulp}{EULP}{End-Use Load Profiles}
\newacronym{ev}{EV}{electric vehicle}
\newacronym{g2v}{G2V}{grid-to-vehicle}
\newacronym{gan}{GAN}{generative adversial network}
\newacronym{gbdt}{GBDT}{gradient-boosted decision trees}
\newacronym{geb}{GEB}{grid-interactive efficient building}
\newacronym{ghg}{GHG}{greenhouse gas}
\newacronym{hdd}{HDD}{heating degree days}
\newacronym{hp}{HP}{heat pump}
\newacronym{hvac}{HVAC}{heating ventilation and air conditioning}
\newacronym{ieq}{IEQ}{indoor environmental quality}
\newacronym{kpi}{KPI}{key performance indicator}
\newacronym{lod}{LoD}{level of detail}
\newacronym{lstm}{LSTM}{long short-term memory}
\newacronym{mape}{MAPE}{mean absolute percentage error}
\newacronym{mdp}{MDP}{markov decision process}
\newacronym{mpc}{MPC}{model predictive control}
\newacronym{occ}{OCC}{occupant-centric control}
\newacronym{ppo}{PPO}{proximal policy optimization}
\newacronym{pv}{PV}{photovoltaic}
\newacronym{rbc}{RBC}{rule-based control}
\newacronym{rc}{RC}{resistance-capacitance}
\newacronym{res}{RES}{renewable energy source}
\newacronym{rlc}{RLC}{reinforcement learning control}
\newacronym{rlcc}{RLC}{resistor, inductor, and capacitor}
\newacronym{rl}{RL}{reinforcement learning}
\newacronym{rmse}{RMSE}{root mean square error}
\newacronym{rnn}{RNN}{recurrent neural network}
\newacronym{sac}{SAC}{soft actor-critic}
\newacronym{saifi}{SAIFI}{System Average Interruption Frequency Index}
\newacronym{soc}{SoC}{state-of-charge}
\newacronym{soo}{SOO}{sequence of operation}
\newacronym{tes}{TES}{thermal energy storage}
\newacronym{tou}{ToU}{time-of-use}
\newacronym{usa}{U.S.}{United States}
\newacronym{v2g}{V2G}{vehicle-to-grid}
\newacronym{vpp}{VPP}{virtual power plant}
\newacronym{wwr}{WWR}{window-to-wall ratio}
\newacronym{zne}{ZNE}{zero-net energy}

\usepackage{subcaption}

\usepackage{algorithm}
\usepackage{algpseudocode}

\usepackage{tabularx}

\usepackage{tabularray}
\UseTblrLibrary{amsmath, counter}
\newcommand{\Eq}{\refstepcounter{equation}(\theequation)}

\usepackage{rotating}

\usepackage{threeparttable}
\usepackage{threeparttablex}

\usepackage{bm}

\usepackage{amssymb}
\usepackage{pifont}

\usepackage{placeins}

\usepackage[noabbrev,capitalise]{cleveref}

\def\tsc#1{\csdef{#1}{\textsc{\lowercase{#1}}\xspace}}
\tsc{WGM}
\tsc{QE}

\usepackage{siunitx}

\begin{document}
\let\WriteBookmarks\relax
\def\floatpagepagefraction{1}
\def\textpagefraction{.001}

\shorttitle{CityLearn v2}    

\shortauthors{Nweye et al.}

\title [mode = title]{CityLearn v2: Energy-flexible, resilient, occupant-centric, and carbon-aware management of grid-interactive communities}   

\author[1]{Kingsley Nweye}[
    orcid=0000-0003-1239-5540
]
\ead{nweye@utexas.edu}
\credit{
Conceptualization,
Methodology,
Software,
Validation,
Formal analysis,
Investigation,
Data curation,
Writing - original draft,
Visualization
}
\affiliation[1]{
    organization={The University of Texas at Austin},
    addressline={301 E. Dean Keeton St., ECJ 4.200}, 
    city={Austin},
    postcode={78712}, 
    state={Texas},
    country={USA}
}

\author[2]{Kathryn Kaspar}[
    orcid=0000-0002-9781-534X
]
\ead{kathryn.kaspar@mail.concordia.ca}
\credit{
Methodology,
Software,
Validation,
Formal analysis,
Investigation,
Data curation,
Writing - original draft
}
\affiliation[2]{
    organization={Concordia University},
    city={Montreal},
    state={Quebec},
    country={Canada}
}

\author[3]{Giacomo Buscemi}[
]
\ead{giacomo.buscemi@polito.it}
\credit{
Methodology,
Software,
Validation,
Formal analysis,
Investigation,
Data curation,
Writing - original draft
}
\affiliation[3]{
    organization={Politecnico di Torino},
    city={Torino},
    country={Italy}
}

\author[4]{Tiago Fonseca}[
    orcid=0000-0002-5592-3107
]
\ead{calof@isep.ipp.pt}
\credit{
Methodology,
Software,
Validation,
Formal analysis,
Investigation,
Data curation,
Writing - original draft,
Visualization
}
\affiliation[4]{
    organization={INESC-TEC/Polytechnic of Porto - School of Engineering},
    city={Porto},
    country={Portugal}
}

\author[5]{Giuseppe Pinto}[
]
\ead{giuseppe-pinto@polito.it}
\credit{
Methodology,
Software,
Investigation,
Data curation,
Writing - review \& editing
}
\affiliation[5]{
    organization={PassiveLogic},
    city={Amsterdam-Centrum},
    state={Noord-Holland},
    country={Nederland}
}

\author[1]{Dipanjan Ghose}[
    orcid=0000-0002-2967-0812
]
\ead{dipanjan02@utexas.edu}
\credit{
Validation,
Writing - original draft,
Visualization
}

\author[1]{Satvik Duddukuru}[
]
\ead{satvik.duddukuru@utexas.edu}
\credit{
Software,
Data curation
}

\author[1]{Pavani Pratapa}[
]
\ead{pavani1404@utexas.edu}
\credit{
Software,
Data curation
}

\author[6]{Han Li}[
]
\ead{hanli@lbl.gov}
\credit{
Methodology,
Writing - review \& editing
}
\affiliation[6]{
    organization={Lawrence Berkeley National Laboratory},
    addressline={1 Cyclotron Rd}, 
    city={Berkeley},
    postcode={94720}, 
    state={California},
    country={USA}
}

\author[1]{Javad Mohammadi}[
]
\ead{javadm@utexas.edu}
\credit{
Writing - review \& editing,
Supervision
}

\author[4]{Luis Lino Ferreira}[
    orcid=0000-0002-5976-8853
]
\ead{llf@isep.ipp.pt}
\credit{
Methodology,
Resources,
Writing - review \& editing,
Supervision
}

\author[6]{Tianzhen Hong}[
    orcid=0000-0003-1886-9137
]
\ead{thong@lbl.gov}
\credit{
Methodology,
Writing - review \& editing,
Supervision
}

\author[2]{Mohamed Ouf}[
]
\ead{mohamed.ouf@concordia.ca}
\credit{
Methodology,
Resources,
Writing - review \& editing,
Supervision
}

\author[3]{Alfonso Capozzoli}[
    orcid=0000-0002-0083-4983
]
\ead{alfonso.capozzoli@polito.it}
\credit{
Methodology,
Resources,
Writing - review \& editing,
Supervision
}

\author[1]{Zoltan Nagy}[
    orcid=0000-0002-6014-3228
]
\ead{nagy@utexas.edu}
\credit{
Conceptualization,
Methodology,
Resources,
Writing - review \& editing,
Supervision,
Project administration
}
\cormark[1]
\cortext[1]{Corresponding author}

\begin{abstract}
As more \acrlongpl{der} become part of the demand-side infrastructure, it is important to quantify the energy flexibility they provide on a community scale, particularly to understand the impact of geographic, climatic, and occupant behavioral differences on their effectiveness, as well as identify the best control strategies to accelerate their real-world adoption. CityLearn provides an environment for benchmarking simple and advanced \acrlong{der} control algorithms including rule-based, model-predictive, and reinforcement learning control. CityLearn v2 presented here extends CityLearn v1 by providing a simulation environment that leverages the End-Use Load Profiles for the U.S. Building Stock dataset to create virtual grid-interactive communities for resilient, multi-agent \acrlongpl{der} and objective control with dynamic occupant feedback. This work details the v2 environment design and provides application examples that utilize \acrlong{rl} to manage \acrlong{bess} charging/discharging cycles, \acrlong{v2g} control, and thermal comfort during heat pump power modulation.
\end{abstract}




\maketitle
\printglossary[type=\acronymtype]
\section{Introduction} \label{sec:introduction}
The electricity grid is undergoing system-wide changes due to the adoption of \glspl{res}, and electrification of buildings as well as transport systems, with the goal of reducing the carbon footprint from power generation and consumption \cite{us_department_of_energy_next-generation_2021}. However, the intermittency of \glspl{res} introduces additional challenges of grid instability due to the mismatch between electricity generation and demand \cite{suberu_energy_2014} and thus, risks reducing the economic value of such sources \cite{gowrisankaran_intermittency_2016}. At the urban scale, this intermittency causes periods of sudden and rapid increase in demand for fossil-fueled generation that threaten grid resiliency. The 'duck curve' is an example of such a period where depleted solar generation resulting from loss of daylight requires a steep ramp up in generation by fossil-fueled power plants to meet demand and is exacerbated by high \gls{pv} penetration \cite{california_independent_system_operator_what_2016}. In extreme cases where power plants are unable to ramp up at the rate needed to meet demand, rotating blackouts are imposed to prevent grid imbalance.

\Glspl{der} in buildings, including \glspl{res}, \glspl{ess}, and heat pumps, can provide the grid with energy flexibility in response to supply deficit, grid signals, changing weather conditions, or occupant preferences \cite{jensen_iea_2017}. \Glspl{ev} also have the capability to provide energy flexibility through \gls{g2v} and \gls{v2g} schemes. Temporary load shedding during on-peak periods, load shifting to off-peak periods, load modulation, and dispatch of generated renewable power for on-site consumption or grid export are several ways in which buildings can use these \glspl{der} to activate their flexibility and reshape their load profile to form a community of grid-interactive buildings \cite{neukomm_grid-interactive_2019}.

As more \glspl{der} become part of the demand-side infrastructure, it is important to quantify the energy flexibility they provide on the urban scale as well as identify best control strategies to accelerate the design and adoption of \gls{dr} programs. Particularly, understanding the impact of geographic, climatic, and occupant behavioral differences on the effectiveness of \glspl{der} in providing flexibility could shape future building design choices and provide a reference for policymakers. However, to carry out such urban scale analysis, an inventory of the current building stock is crucial. The \gls{eulp} for the \gls{usa} Building Stock dataset \cite{wilson_end-use_2022}, generated with ResStock \cite{wilson_resstock_2017} and Comstock \cite{horsey_comstock_2020} engines, uses physics-based energy models to provide over 900,000 synthetic building models that are calibrated with real-world data and represent the residential and commercial building stock in the \gls{usa} The buildings differ in end-use fuel source, occupant behavior that affect energy profiles, archetype, construction material, climatic region and other distinguishable building characteristics. With the availability of such a large and diverse energy model repository, residential, commercial, and mixed-use grid-interactive communities to study the impact of \glspl{der} and \gls{dr} programs are made possible.

It is, however, challenging to coordinate multiple \glspl{der} in a single building or multiple buildings while ensuring efficient and flexible operation that does not increase the risk of occupant discomfort. These \glspl{der} could also cause a deviation from the \gls{bau} load profile and could introduce new peaks \cite{lee_unintended_2022,nweye_merlin_2023}. Also, \gls{ev} loads, if not properly managed, have the potential to destabilize the grid with up to 20\% increase in peak electricity demand \cite{birk_jones_impact_2022} and 35\% increase in electricity consumption \cite{galvin_are_2022}.

Aside the risk of power outages caused by supply-demand mismatch, wildfires \cite{zanocco_when_2021} and extreme weather such as heat waves \cite{feng_tropical_2022}, and winter storms \cite{busby_cascading_2021} also exacerbate the likelihood of power outages leading to unserved energy and occupant discomfort. Thus, a control solution for \glspl{der} must be resilient during periods of curtailed supply or extreme cases of blackouts to mitigate the disaster vulnerability of occupants.

Furthermore, the stochasticity of occupant behavior may result in different energy profiles, thus indicates the influence that appliance usage, thermostat setpoints, and other occupant-driven end-uses have on energy consumption \cite{akbari_occupancy_2021}. However, typical thermal comfort models such as Fanger's Predicted Mean Vote-Predicted Percentage Dissatisfied model \cite{fanger_thermal_1970} do not capture occupant preferences and behaviours that are observed in residential buildings. Smart thermostat data, such as ecobee's Donate Your Data dataset \cite{luo_ecobee_2022} can be leveraged to develop more realistic representations of both thermostat setpoint profiles and thermostat override behavior by occupants in the residential setting. In this way, thermostat setpoints can be automatically controlled for the implementation of \gls{dr}, while occupants can choose to override the \gls{dr} changes due to discomfort, thus balancing both energy efficiency and thermal comfort.

Compared to simpler \gls{rbc} \cite{lu_benchmarking_2023}, advanced control algorithms such as \gls{mpc} \cite{drgona_all_2020} and \gls{rlc} \cite{nagy_ten_2023-1} could provide solutions to these control challenges by adapting their control policy to unique building characteristics, occupant behaviors, and pricing or demand response signals from the power grid, while cooperating towards improving multi-objective \glspl{kpi}. When improved, these \gls{kpi} such as minimizing the peak load in a building district and maintaining building-specific comfort needs, ensure energy efficiency and high \gls{ieq} in buildings.

A research gap in the adoption of advanced control for district or community-scale \gls{der} control and \gls{dr} is the ability to benchmark algorithm performance in buildings \cite{vazquez-canteli_reinforcement_2019}. Physics-based models accurately capture the thermal dynamics in buildings and are the industry standard for assessing the impact of control in the built environment. However, they require extensive domain knowledge, input data and attention to detail to model \glspl{bes} as well as significant effort to integrate standardized third-party control libraries for co-simulation and control algorithm benchmarking. Building emulators provide various levels of abstractions of \glspl{bes}, which enables the designer to focus on the control implementation. Such emulators include Energym \cite{scharnhorst_energym_2021} and BOPTEST \cite{blum_building_2021}, which provide high-fidelity \gls{bes} models. Although these emulators take advantage of robust simulation engines including EnergyPlus and Modelica-based Building Library, their dependency on such engines raises the entry level for users. Moreover, they are designed for specific \gls{bes} or building environments, do not consider the growing influence of \glspl{ev} on building loads, do not model grid-resiliency nor occupant behavior, and do not allow for district-level control nor \gls{kpi} evaluation. The DOPTEST framework \cite{arroyo_prototyping_2023} that builds upon BOPTEST to support district-level and multi-agent control requires compute-intensive co-simulation and is limited to building and district environments provided within the framework. OCHRE \cite{blonsky_ochre_2021} is a Python-based emulator for both single and multi building control that makes use of a detailed RC thermal dynamics model with user-configurable envelope parameters and \glspl{bes}. Its \gls{bes} models are defined similarly to EnergyPlus \glspl{bes} thus, provides a blend of high fidelity modeling and custom buildings without the need for co-simulation with EnergyPlus. However, its interface does not follow any standardized environment protocol such as Gymanasium \cite{towers_gymnasium_2023} neither is it directly compatible with standardized control libraries e.g. Stable-Baselines3 \cite{raffin_stable-baselines3_2021} and RLlib \cite{liang_rllib_2017} for use in control algorithm benchmarking.

CityLearn is an open-source Gymnasium environment for the easy implementation and benchmarking of \gls{rbc}, \gls{rlc} and \gls{mpc} algorithms for \glspl{der} in a \gls{dr} setting. CityLearn is used to reshape the aggregated electricity load profile by controlling \glspl{der} in a district of diverse buildings, and allows for multi-agent control and district-level \glspl{kpi} evaluation. Introduced in \cite{vazquez-canteli_citylearn_2019}, it has been applied extensively for \gls{der} control benchmarking in scenarios of \gls{dr} \cite{kathirgamanathan_centralised_2020,dhamankar_benchmarking_2020,deltetto_exploring_2021,glatt_collaborative_2021,pinto_coordinated_2021, qin_neorl_2021,pinto_enhancing_2022}, voltage regulation \cite{pigott_gridlearn_2022} as well as control policy meta-learning \cite{zhang_metaems_2022}, and transfer learning \cite{nweye_merlin_2023}.

The CityLearn v2 release that we present here provides a framework that incorporates thermal dynamics modeling in \cite{pinto_data-driven_2021} that allows for heat pump capacity/power control to provide partial load satisfaction, precooling and preheating services in buildings. Additionally, this new release integrates an \gls{ev} module to provide \gls{g2v} and \gls{v2g} control, as well as occupant modeling for thermostat setpoint override to simulate realistic \gls{occ} and thermal comfort assessment. CityLearn v2 also provides an integration with the \gls{eulp} for the \gls{usa} Building Stock dataset based on the work by \citeauthor{bs2023_1404} \cite{bs2023_1404} as well as the functionality to assess the resiliency of control algorithms during power outage events through a stochastic power outage model based on distribution system reliability metrics \cite{IEEEGuideElectric2012} such that it is generalizable by providing custom or location-specific metrics. For compatibility with standard control interfaces, the v2 environment follows the latest Gymansium control protocol and provides wrappers for interfacing with Stable-Baselines3 and RLlib single and multi-agent algorithms. Thus, we summarize the contributions of our work as providing:

\begin{enumerate}
    \item a simulation environment that leverages realistic building-stock datasets for resilient multi-agent, \gls{der} and objective control with dynamic occupant feedback in grid-interactive communities;
    \item an all-in-one environment without the need for co-simulation that is designed according to the standardized Gymnasium interface; and
    \item extensible interface for standardized control algorithm libraries.
\end{enumerate}

The remainder of our paper is organized as follows: \cref{sec:environment} explains the environment and its internal models in detail, while \cref{sec:control} describes the environment's control interface and configurations. We provide example applications of the CityLearn v2 environment in \cref{sec:examples} and discuss the implications of the new environment in building controls research as well as its limitations in \cref{sec:discussion}. Finally, we conclude in \cref{sec:conclusion}. An \nameref{sec:appendix} section provides supporting environment documentation.
\section{Environment} \label{sec:environment}
\begin{figure*}
    \centering
    \includegraphics[width=1.0\textwidth]{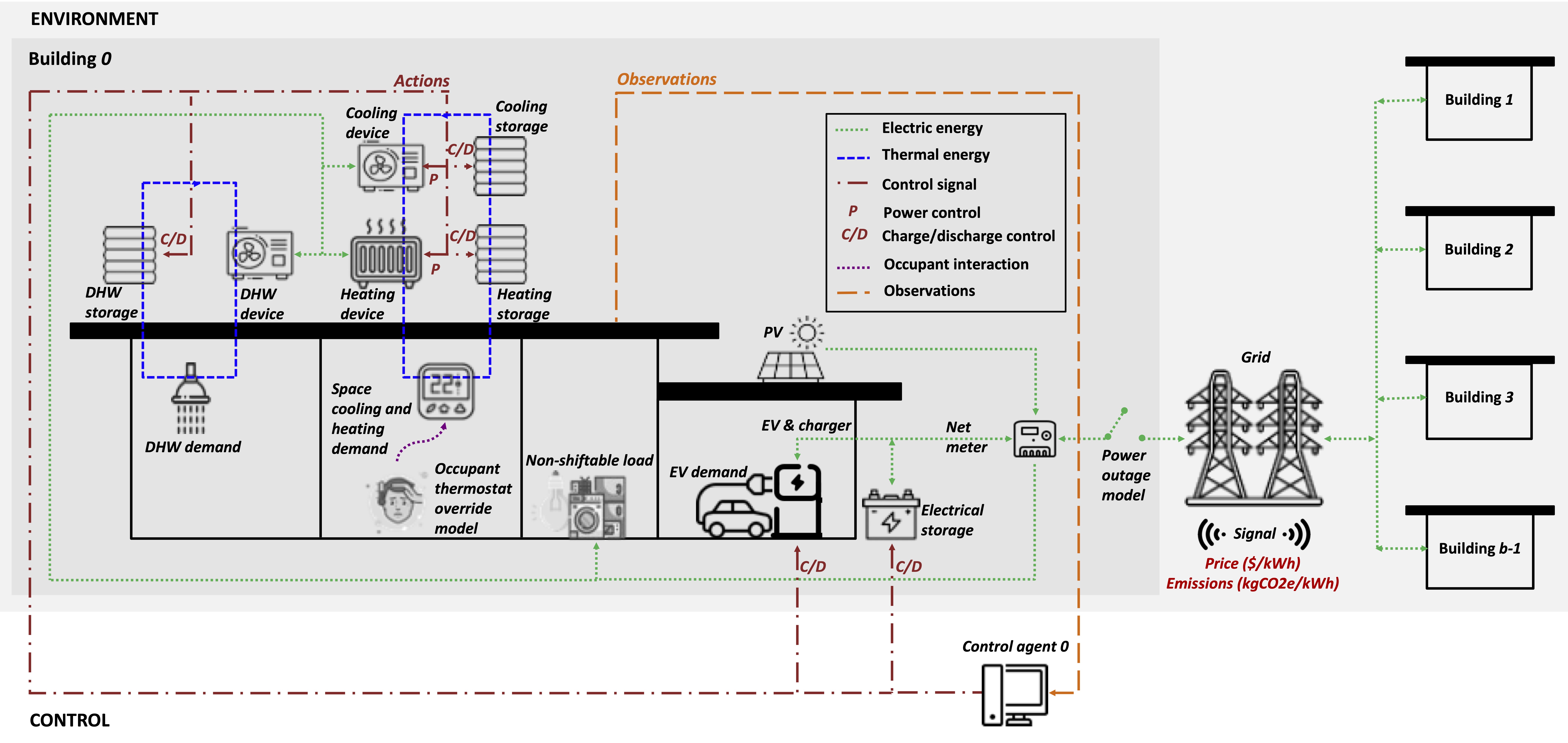}
    \caption{CityLearn building model including electricity sources that power controllable \glspl{der} including electric devices and \glspl{ess}, used to satisfy thermal and electrical loads as well as provide the grid with energy flexibility. A distinction is made between environment and control aspects of a building to show the transfer of actions from the control agent and reception of measurable observations by the control agent that quantifies the building's states.}
    \label{fig:citylearn_building_schematic}
\end{figure*}

CityLearn\footnote{ANONYMOUS} models a district of buildings with similar or different loads, electric devices, \glspl{ess} and electricity sources that satisfy the loads as shown in \cref{fig:citylearn_building_schematic}. There is no upper limit on the number of buildings in a district and a district can have as few as one building. The electric devices and \glspl{ess} are one of many \glspl{der} described in \cref{sec:environment-distributed_energy_resources}. We further describe how these \glspl{der} are used in a building to satisfy loads, store energy, or provide electricity supply in \cref{sec:building_model}. The power outage model is described in \cref{sec:environment-power_outage_model}, while \cref{sec:key_performance_indicators,sec:datasets} define the in-built \glspl{kpi} and datasets for control benchmarking. We end this section in \cref{sec:framework_for_representative_neighborhoods} by describing a framework that leverages the \gls{eulp} dataset to create virtual grid-interactive communities for CityLearn.

\subsection{Building model} \label{sec:building_model}
The building model is illustrated in \cref{fig:citylearn_building_schematic} where we show the transfer of actions from the control agent and its reception of measurable observations that describe the building's states. This building-control interface is explained in \cref{sec:control}.

A building is a combination of electricity sources that power controllable \glspl{der}, including electric devices and \glspl{ess} used to satisfy thermal and electrical loads as well as provide the grid with energy flexibility (\cref{sec:environment-building-loads,sec:environment-building_model-heating_ventilation_and_air_conditioning,sec:environment-building_model-energy_storage_systems,sec:environment-building_model-electricity_sources}). The building is modeled as a single thermal zone where space thermal loads affect its indoor dry-bulb temperature. We use a data-driven thermal dynamics model to quantify the effect of the thermal load on temperature (\cref{sec:building-thermal_dynamics_model}) and an occupant model that has the ability to override the temperature setpoint is described in \cref{sec:environment:occupant_model}.

\subsubsection{Loads} \label{sec:environment-building-loads}
There are five loads in a building including space cooling, space heating, \gls{dhw} heating, electric equipment, and \gls{ev} loads. The space cooling and heating loads refer to the energy needed to maintain the indoor dry-bulb temperature at its setpoint. We make no distinction between cooling and heating setpoints and CityLearn does not allow for simultaneous space cooling and heating in the same time step for a building. The \gls{dhw} heating load is the total heating energy needed to satisfy hot water end-uses such as shower, bathroom, and kitchen sinks, and other water end uses requiring water heating that are not space heating-related. Electric equipment refer to non-shiftable plug loads such as lighting, entertainment and kitchen appliances. The \gls{ev} load is the energy required to charge an \gls{ev} to a scheduled departure \gls{soc}. Within CityLearn, the space cooling, space heating, \gls{dhw} heating, electric equipment, and \gls{ev} loads are called \verb|cooling_demand|, \verb|heating_demand|, \verb|dhw_demand|, \verb|non_shiftable_load|, and \verb|ev_demand|.

Not all loads need to exist in a building e.g., a building situated in a heating dominant climate may not have cooling loads year-round. Also, anyone or all of these loads are either known a priori from \gls{beps} \cite{vazquez-canteli_marlisa_2020} or real-world measurement \cite{nweye_merlin_2023}. In these instances, the ideal load must be satisfied. Alternatively, they are controlled loads and are inferred at runtime e.g., heat pump power control driving space cooling or heating loads. To satisfy these loads in either the ideal or control-action case, we make use of \gls{hvac} systems directly (\cref{sec:environment-building_model-heating_ventilation_and_air_conditioning}), or \glspl{ess} (\cref{sec:environment-building_model-energy_storage_systems}) through load shifting, where these systems are powered by the electricity sources described in \cref{sec:environment-building_model-electricity_sources}.

\subsubsection{Electric devices} \label{sec:environment-building_model-heating_ventilation_and_air_conditioning}
The \verb|cooling_device|, \verb|heating_device|, and \verb|dhw_device| are \gls{hvac} electric device objects in CityLearn that are used to satisfy the space cooling, space heating and \gls{dhw} heating loads respectively. The \verb|cooling_device| is a heat pump while the \verb|heating_device| and \verb|dhw_device| are either heat pump or electric heater type. The CityLearn version at the time of writing allows for \verb|cooling_device| and \verb|heating_device| control subsequently, driving \verb|cooling_demand| and \verb|heating_demand|.

These \gls{hvac} systems are powered directly by one or more of the electricity sources in \cref{sec:environment-building_model-electricity_sources} and if available, may be used to charge \gls{tes} systems in the building. 

The \verb|electric_vehicle_charger| is used to satisfy \gls{ev} load or acts an intermediate link between the \gls{ev} and grid for \gls{v2g} applications.

\subsubsection{Energy storage systems} \label{sec:environment-building_model-energy_storage_systems}
There are up to five optional and controlled \glspl{ess} in a building including \verb|cooling_storage|, \verb|heating_storage|, \verb|dhw_storage|, \verb|electrical_storage|, and \verb|electric_vehicle| \gls{ess} objects. The \verb|cooling_storage|, \verb|heating_storage|, and \verb|dhw_storage| are \gls{tes} \gls{der} type and provide space cooling, space heating and \gls{dhw} heating load shifting flexibility respectively. They are charged by the \gls{hvac} device used to meet the thermal load which they service e.g. is the \verb|cooling_device| charging the \verb|cooling_storage|. However, the building's thermal load is first satisfied and what is left of the \gls{hvac} device's nominal power is used to charge the \gls{tes}. When discharging, the energy from the \gls{tes} is first used to meet the building's thermal load before the \gls{hvac} device.

The \verb|electrical_storage| is a \gls{bess} \gls{der} type that powers any of the aforementioned electric devices when in discharge mode or is powered by one or more of the electricity sources in \cref{sec:environment-building_model-electricity_sources}. If the \verb|electrical_storage| discharges more energy than needed to meet the building loads, the excess is sent to the grid as part of the building's net export.

The \verb|electric_vehicle| is an \gls{ev} \gls{der} type and performs similar function as the \verb|electrical_storage|, however, the \gls{ev} is available on a schedule defined by its arrival and departure times. The \gls{ev} can be used in three modes: \gls{g2v}, \gls{v2g}, and no control (i.e., where the \gls{ev} acts as a load without any possible control over its charging) \cite{fonseca_evlearn_2024}.

\subsubsection{Electricity sources} \label{sec:environment-building_model-electricity_sources}
The electric devices are primarily powered by the electric grid. CityLearn at the time of writing, does not have a grid model so the power a building is able to draw from the grid at a given time step is unconstrained, except in the case of a power outage (\cref{sec:environment-power_outage_model}). Optionally, a building may have a \gls{pv} system (\verb|pv|) that provides self-generation as a first source of electricity before the grid. Optional \verb|electrical_storage| and \verb|electric_vehicle| are charged by the grid and \verb|pv| but also augment the \verb|pv| and grid when in discharge mode to supply the building with electricity. Excess self-generation, \verb|electrical_storage| and \verb|electric_vehicle| discharge are sent to the grid as apart of the building's net export.

\begin{multline}
E_t^{\textrm{building, net}} = E_t^{\textrm{cooling\_device}} + E_t^{\textrm{heating\_device}} \\ + E_t^{\textrm{dhw\_device}} + E_t^{\textrm{non\_shiftable\_load}} + E_t^{\textrm{electrical\_storage}} + \\ E_t^{\textrm{electric\_vehicle\_charger}} + E_t^{\textrm{pv}}
    \label{eqn:building-net_electricity_consumption}
\end{multline}

Given the aforementioned electric devices, \glspl{ess} and electricity sources, the net electricity consumption of the building (\cref{eqn:building-net_electricity_consumption}) is thus, the sum of the positive electricity consumption by the \verb|cooling_device| ($E_t^{\textrm{cooling\_device}}$), \verb|heating_device| ($E_t^{\textrm{heating\_device}}$), \verb|dhw_device| ($E_t^{\textrm{dhw\_device}}$), and \verb|non_shiftable_load| ($E_t^{\textrm{non\_shiftable\_load}}$), mixed polarity electricity consumption of the \verb|electrical_storage| ($E_t^{\textrm{electrical\_storage}}$), and \verb|electric\_vehicle\_charger| ($E_t^{\textrm{electric\_vehicle\_charger}}$), and negative polarity electricity generation by the \verb|pv| ($E_t^{\textrm{pv}}$). $E_t^{\textrm{cooling\_device}}$, $E_t^{\textrm{heating\_device}}$, and $E_t^{\textrm{dhw\_device}}$ are the electricity to be used to satisfy the \verb|cooling_demand|, \verb|heating_demand|, and \verb|dhw_demand| after accounting for energy discharge from the \verb|cooling_storage|, \verb|heating_storage|, and \verb|dhw_storage|. Therefore, a case where there the control action for these three \glspl{tes} is adequate to fully satisfy the loads, $E_t^{\textrm{cooling\_device}}$, $E_t^{\textrm{heating\_device}}$, and $E_t^{\textrm{dhw\_device}}$ will equal zero.

The district-level net electricity consumption is the sum of $E_t^{\textrm{building, net}}$ for all buildings as defined in \cref{eqn:district-net_electricity_consumption}. 

\begin{equation}
    E_t^{\textrm{district, net}} = \sum_i^{b-1}{E_t^{\textrm{building }i, \textrm{net}}}
    \label{eqn:district-net_electricity_consumption}
\end{equation}

\subsubsection{Thermal dynamics model} \label{sec:building-thermal_dynamics_model}
The building's thermal dynamics is modeled as a \gls{lstm} network that predicts the indoor dry-bulb temperature, $T_t^{\textrm{in}}$, given a lookback, $l$, of previous observations in the building \cite{pinto_data-driven_2021}. The model function is defined in \cref{eqn:thermal_dynamics} where its independent variables include indoor dry-bulb temperature ($T^{\textrm{in}}$), outdoor dry-bulb temperature ($T^{\textrm{out}}$), cooling or heating load depending on \verb|hvac_mode| defined in \cref{sec:control-observation_space} ($Q^{\textrm{cooling | heating}}$), direct solar irradiance ($Q^{\textrm{solar, direct}}$), diffuse solar irradiance ($Q^{\textrm{solar, diffuse}}$), occupant count ($U$), month ($m$), day-of-week ($d$), and hour ($h$). While the $T^{\textrm{in}}$ input to the model is the previous $l$ time steps, other input are their observations between the current time step, $t$, and previous $l-1$ time steps. \Cref{sec:environment-temperature_dynamics_model_methodology} explains the methodology for generating the model fitting data and describes how these data are used in training and testing the \gls{lstm} network.

\begin{multline}
    T_t^{\textrm{in}} = f\Big(T_{t-1 \dots t - l}^{\textrm{in}}, \Big(T^{\textrm{out}}, Q^{\textrm{cooling | heating}}, \\ Q^{\textrm{solar, direct}}, Q^{\textrm{solar, diffuse}}, U, m, d, h \Big)_{t \dots t - (l - 1)}\Big)
    \label{eqn:thermal_dynamics}
\end{multline}

\subsubsection{Occupant-thermostat override model} \label{sec:environment:occupant_model}
Local or regional thermostat data including indoor air temperature, date/time, HVAC system state (e.g. heat, cool) and thermostat events (e.g. override occurrence) can be used to develop occupant-thermostat override models. Smart thermostats are programmable and allow for the user to set various program settings during specified hours and duration, such as `Home', `Away', and `Sleep'. When a user decides to change from the set program, an override occurs, and generally this behavior can be interpreted as discomfort-driven. However, not all occupants have the same thermostat setpoint preferences and subsequent override behavior. As such, to develop the occupant-thermostat override models, we firstly (a) cluster the thermostat setpoint profiles to identify distinct thermostat setpoint preferences in the region; (b) develop a model to predict the probability of a setpoint override for each occupant type; (c) estimate the magnitude of the thermostat setpoint override.

Firstly, we develop models for three periods (summer, winter, shoulders) separately, as HVAC system state and occupant override behavior are different during each period. Next, we query the setpoint data for one period (e.g. winter) and extract discomfort-driven behavior from the dataset. In the case study presented in \cite{liu_modeling_2021}, seven out of 20 occupants showed no correlation between the occurrence of a setpoint increase and any environmental variable, e.g. indoor or outdoor air temperature, thus showing the difficulties in interpreting and using thermostat data. In a winter heating scenario, we expect an inverse relationship between the indoor air temperature and a setpoint override (the data was organized such that `0' represented no override and `1' represented an override). We thus firstly used Pearson Correlation to extract all homes/occupant data with a correlation less than -0.20 (some correlation) between the indoor air temperature and an increase to the temperature setpoint, thus focusing the modeling on this subset of the data.

Using the remaining data, we then used k-means correlation to identify clusters based on the average indoor air temperature during a setpoint override, separating the overrides by whether they increased or decreased the setpoint. The Silhouette score \cite{rousseeuw_silhouettes_1987} was used to determine the number of clusters, and we visualized the average daily setpoint profile as well as a histogram of the indoor air tempterature to ensure unique clustering. It would also be possible to do k-shape clustering on the average daily setpoint profiles as well to create the unique clusters, as was done in \cite{panchabikesan_investigating_2021}.

We next model the probability of a setpoint override for each occupant type, again separating setpoint increase and setpoint decrease behavior for the overrides. If occupancy information is available, extract only the time steps in which the home is occupied for this modeling; otherwise, assume an acceptable occupancy rate for the dataset. We then replicate the methodology used in \cite{liu_modeling_2021,gunay_development_2018} and use discrete-time Markov logistic regression models to predict the probability of a thermostat setpoint override, an industry standard as referenced in \cite{doca_critical_2019}. Equation \ref{eqn:override} shows the form of the logistic regression equation, with the indoor air temperature, $T_{in}$, representing the input variable and \textit{p} representing the probability of a setpoint increase or decrease (override = 1, in either case, but the scenarios are evaluated separately). Coefficients \textit{a} and \textit{b} are determined in the training, and the fit of the curve is evaluated using the p-value.

\begin{equation}\label{eqn:override}
p (override=1) = \frac{1}{1 + e^{-(a + bT_{\text{in}})}}
\end{equation}

Once the probability of an override is modeled, the next step is determining by how much the setpoint is changed in the event that an override occurs. Depending on the data available, it is possible to randomly sample from the distribution of the setpoint overrides to estimate the change in setpoint during the override. We further developed random forest models to classify the change to the setpoint during an override as either small (less than \SI{0.5}{\degreeCelsius}) or large (greater than \SI{0.5}{\degreeCelsius}).

\subsection{Power outage model} \label{sec:environment-power_outage_model}
A power outage is defined as a time series, $O$, of binary signals where $O_t=0$ indicates no outage at time step, $t$, and $O_t=1$ indicates an outage. A sequence of consecutive values of $O_t=1$ is regarded as a power outage event and overlapping events are merged into one event. These signals are either arbitrarily defined or generated using some stochastic model.

We provide a stochastic power outage model based on distribution system reliability metrics \cite{IEEEGuideElectric2012} including the number of non-momentary electric interruptions, per year, an average customer experiences (\gls{saifi}), and the average number of minutes it takes to restore non-momentary electric interruptions (\gls{caidi}). These metrics are reported annually and location-specific \cite{us_energy_information_administration_table_nodate} allowing for model adaptation to real-world locations and scenarios. A given day of the year having an outage event or not is determined by randomly sampling from the Bernoulli distribution in \cref{eqn:power_outage-probability_day_is_outage}, where the probability of an event, $p$, is the ratio of \gls{saifi} to number of days in a non-leap year (365). The start of an event is randomly sampled from a uniform distribution, where \cref{eqn:power_outage-start_time_step} is an example of such distribution for an hourly time step. Finally, the duration of each power outage event is sampled from an exponential distribution with rate, $\lambda$, set to \gls{caidi} (\cref{eqn:power_outage-duration}). Sampling from these distributions for a fixed \gls{saifi} and \gls{caidi} is controlled by the model's random seed.

\begin{equation}
     \textrm{Outage event} \sim \textrm{Bernoulli}\Big(p=\frac{\textrm{SAIFI}}{365}\Big)
     \label{eqn:power_outage-probability_day_is_outage}
\end{equation}

\begin{equation}
    \textrm{Outage start hour} \sim U(0, 23)
    \label{eqn:power_outage-start_time_step}
\end{equation}

\begin{equation}
    \textrm{Outage duration (minutes)} \sim Exp(\lambda=\textrm{CAIDI})
    \label{eqn:power_outage-duration}
\end{equation}

\begin{equation}
    E_t^{\textrm{available}} = \begin{cases}
            E_t^{\textrm{pv}} \\ - \Big(
        E_t^{\textrm{cooling\_device}}
        + E_t^{\textrm{heating\_device}}
        \\ + E_t^{\textrm{dhw\_device}}
        + E_t^{\textrm{non\_shiftable\_load}}
        \\ + E_t^{\textrm{electrical\_storage}}
        \\ + E_t^{\textrm{electric\_vehicle\_charger}}
    \Big) & \text{ if } O_t > 0 \\
            \infty & \text{ otherwise}
        \end{cases}
        \label{eqn:power_outage_flexibility}
\end{equation}

\Cref{eqn:power_outage_flexibility} defines the available electricity supply during a power outage event and normal grid operation. In an outage event, the grid is unable to provide buildings with electricity and buildings can only make use of the flexibility provided by \glspl{ess} and \gls{pv} system after subtracting electricity use by (1) the \verb|cooling_device|, \verb|heating_device|, and \verb|dhw_device| to meet thermal loads and charge \glspl{tes}, (2) \verb|non_shiftable_load|, as well as (3) \verb|electrical_storage| and \verb|electric_vehicle_charger| if in charging mode. Whereas, during normal operation, there is unlimited supply from the grid.

\subsection{Key performance indicators} \label{sec:key_performance_indicators}
\Cref{tab:kpi} summarizes the \glspl{kpi} within CityLearn that are used for thermal-comfort, energy cost, environmental, energy, and resilience evaluation. These \glspl{kpi} are calculated at either the building or district level by default and \glspl{kpi} defined at the building level are evaluated at the district level by calculating the building-level average. The \glspl{kpi} are defined such that the control objective is to minimize their value.

\subsection{Datasets} \label{sec:datasets}
\Cref{tab:citylearn_datasets} is a summary of the in-built datasets in CityLearn that were designed for a distribution of control problems including those specific to The CityLearn Challenge \cite{vazquez-canteliCityLearnChallenge2020-2,nagy_citylearn_2021,nweye_citylearn_2022}. Each dataset is a unique environment configuration where environments differ by their location and climate zone that influence their weather file and building characteristics, time series date range, building count, model availability (thermal, occupant and power outage models) that affect the complexity of the control problem, control signal availability including electricity pricing and carbon emissions time series, building loads, controllable \glspl{ess}, and \gls{pv} system availability for self-generation.

\subsection{Environment design workflow} \label{sec:framework_for_representative_neighborhoods}
\begin{figure*}
    \centering
    \includegraphics[width=1.0\textwidth]{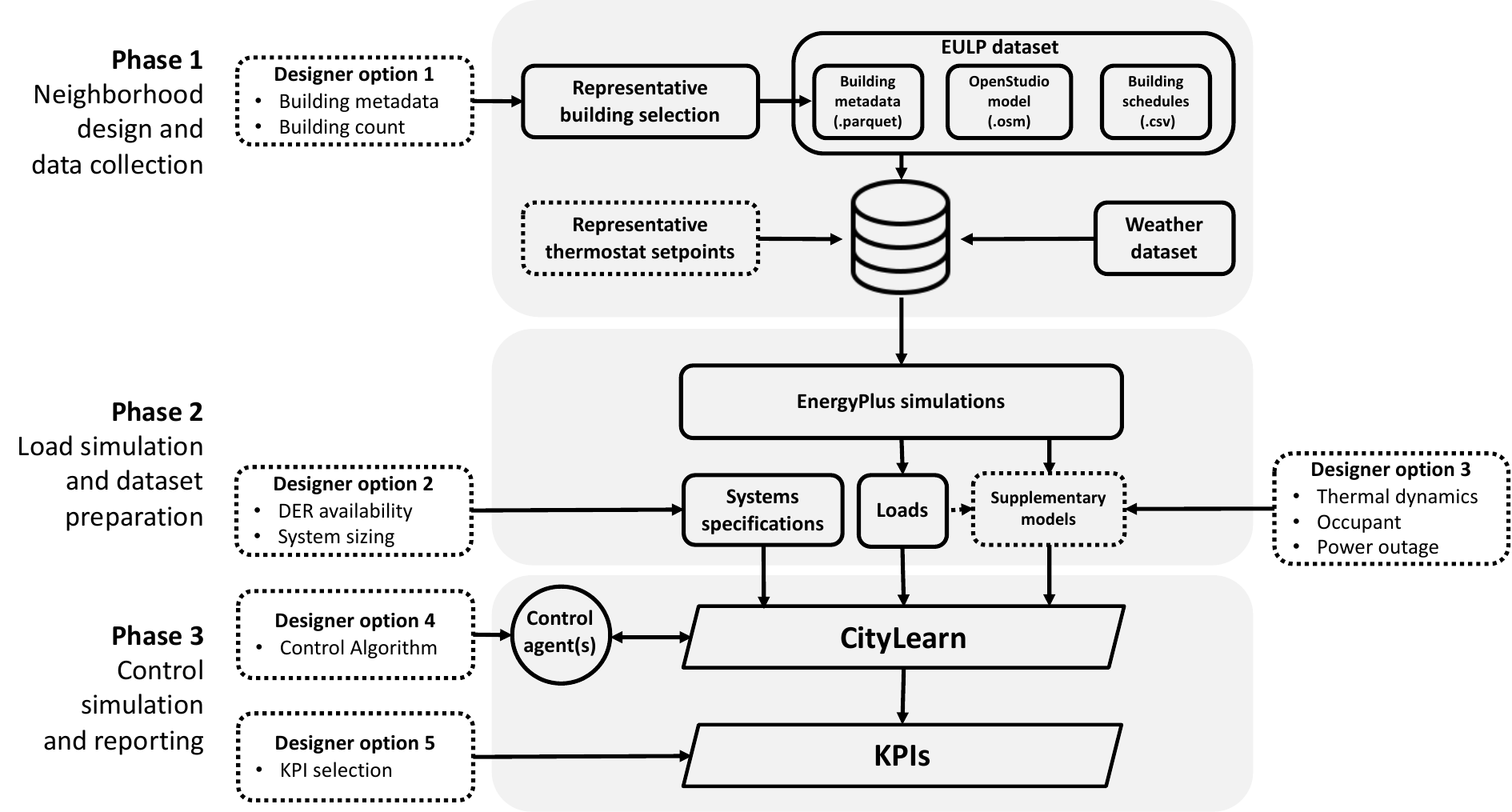}
    \caption{Framework for designing virtual grid-interactive communities in CityLearn from the \acrlong{eulp} for the United States Building Stock dataset (adapted from \cite{bs2023_1404}).}
    \label{fig:eulp_integration}
\end{figure*}

\Cref{fig:eulp_integration} depicts a framework for designing virtual grid-interactive communities in CityLearn by leveraging the \gls{eulp} dataset as well as other data sources \cite{bs2023_1404}. This framework is split into three phases: (1) neighborhood design and data collection, (2) load simulation and dataset preparation and (3) control simulation and reporting. 

In the neighborhood design and data collection phase, the neighborhood designer pre-filters the \gls{eulp} dataset for buildings whose metadata match certain criteria such as archetype, vintage, location, and \gls{hvac} equipment type and the designer may choose to define the building count in the neighborhood. Random selection or data-driven methods are used to select buildings up to the specified count. A data-driven approach is provided in \cite{bs2023_1404} that makes use of cluster-frequency-based sampling from buildings clustered by metadata similarities including orientation, decade of construction, number of occupants, infiltration rate, ceiling, slab and wall insulation, \gls{eui} and \gls{wwr}. The building metadata, occupancy and load schedules, weather data and energy models for the selected buildings are then stored in a central database for easy retrieval and manipulation. Optionally, the variance in building loads and indoor environmental conditions across buildings is improved by replacing the default thermostat setpoint schedules with real-world thermostat setpoints from the ecobee Donate Your Data dataset \cite{luo_ecobee_2022}.

In the building load simulation and dataset preparation phase, the collected data are used to run \gls{beps} simulations following the methodology for thermal dynamics modeling in \cref{sec:building-thermal_dynamics_model}. The \gls{lstm} thermal dynamics model is interchangeable with other thermal dynamics modeling methodology e.g. \gls{rc} modeling. In addition to the data needed to train the thermal dynamics model other ideal loads data including \gls{dhw} heating loads and plug loads are retrieved from the \gls{beps}. The choice of occupant and power outage models are also decided in this second phase. Finally in the second phase, designer input for \gls{der} availability as well as their technical specifications are defined. 

The simulated loads, models and system specifications from the second phase are utilized to create a virtual representation of the intended neighborhood in the control simulation and reporting phase. The designer selects a control algorithm to manage the \glspl{der}. Post-simulation evaluation of control performance is achieved by the user-selected \glspl{kpi}.
\section{Control} \label{sec:control}
\begin{figure*}[]
    \centering
    \begin{subfigure}[t]{0.3\textwidth}
        \centering
        \includegraphics[width=\textwidth]{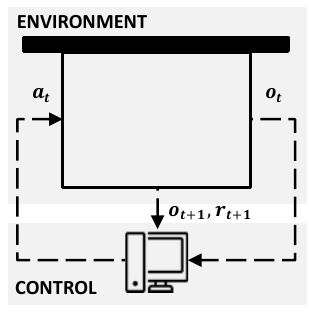}
        \caption{Farama Foundation Gymnasium interface.}
        \label{fig:control_summary-gym_interface}
    \end{subfigure}
    \hspace{0.01\textwidth}
    \begin{subfigure}[t]{0.67\textwidth}
        \centering
        \includegraphics[width=\textwidth]{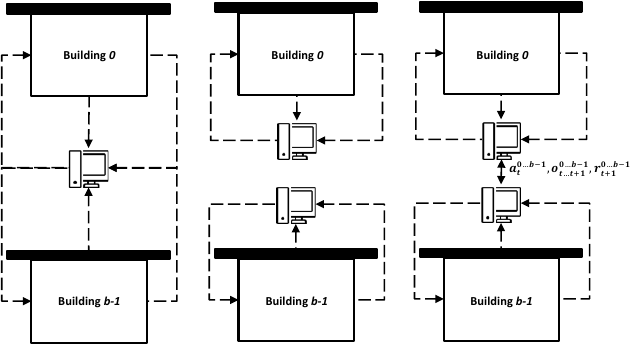}
        \caption{Single-agent (left), independent multi-agent (middle), and coordinated multi-agent (right) control configurations.}
        \label{fig:control_summary-control_architecture}
    \end{subfigure}
    \caption{Control summary in CityLearn.}
    \label{fig:control_summary}
\end{figure*}

Control in CityLearn is defined by its interface and configuration as shown in \cref{fig:control_summary}. The CityLearn environment makes use of the Farama Foundation Gymnasium interface \cite{towers_gymnasium_2023}, built upon the OpenAI Gym interface \cite{brockman_openai_2016}, for standardized \gls{rlc} environment design. \Cref{fig:control_summary-gym_interface} illustrates this interface where there is an observation-action-reward exchange loop between the environment and control agent as the environment transitions from one time step to another. In the current time step, $t$, the control agent receives the environment's observations, $o_t$ and prescribes actions $a_t$. The actions are applied to the environment to affect the observations at the next time step, $o_{t+1}$. $o_{t+1}$ and a reward, $r_{t + 1}$ (from reward function, $R$) that quantifies the quality of $a_t$ in optimizing the outcome of a control objective or \gls{kpi} are returned to the control agent to teach it to learn a control policy, $\pi$. $\pi$ maps actions to observations that maximize the cumulative reward over an episode i.e., the terminal state of the environment, after initialization ($t=0$), beyond which there are new observations. \Cref{sec:control-observation_space,sec:control-action_space} describe the observation and action spaces in CityLearn while \cref{sec:control-reinforcement_learning_control_reward_functions} outlines the internally defined reward functions in the environment.

\Cref{fig:control_summary-control_architecture} shows three possible control configurations in CityLearn namely; single-agent, independent multi-agent, and coordinated multi-agent. In the single-agent configuration, there is a one-to-many relationship between the control agent and buildings where a centralized agent collects observations and prescribes actions for all \glspl{der} in the district and, receives a single reward value each time step to learn a generalized control policy. This is akin to an energy aggregator controlling flexible resources in a distributed manner. The independent multi-agent configuration has a one-to-one agent-building relationship thus, there are as many rewards as buildings each time step and a unique control policy is learned for each building. The coordinated multi-agent configuration is similar to the independent multi-agent configuration except that agents can share information to achieve cooperative objectives e.g. district peak reduction or competitive objectives e.g. price bidding in the energy flexibility market. \Cref{sec:control-control_algorithms} discusses some of the in-built control agent algorithms in CityLearn that use one or more of these configurations as well as CityLearn's provisions for interfacing with third-party standardized control algorithm libraries. 

\begin{figure*}
    \centering
    \includegraphics[width=1.0\textwidth]{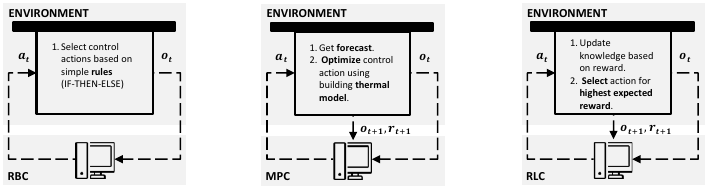}
    \caption{Integration of \acrfull{rbc}, \acrfull{mpc}, and \acrfull{rlc} with CityLearn.}
    \label{fig:control_theories}
\end{figure*}

We emphasize that CityLearn is not limited to \gls{rlc} algorithms alone despite its Gymnasium interface as it works with other simple control theory algorithms e.g., \gls{rbc} \cite{nweye_real-world_2022} as well as advanced control theory algorithms e.g., \gls{mpc} \cite{zhan_comparing_2023} as shown in \cref{fig:control_theories}. In \gls{rbc}, the reward is not utilized while in \gls{mpc}, the reward is akin to the control objective. Also, \gls{rbc} policy is static and does not consider the observations in the following time step to update its rules.
\section{Examples} \label{sec:examples}
This section provides example applications of the CityLearn v2 environment. \Cref{sec:examples-control_task_distribution} showcases 17 different control tasks of differing complexity, \cref{sec:examples-ess_control_in_neighborhoods} applies the \gls{eulp} dataset to create three representative neighborhoods for \gls{ess} control, \cref{sec:examples-vehicle_to_grid} provides a commercial building district \gls{v2g} application for a European \gls{ev} dataset, and \cref{sec:examples-occ_in_control} demonstrates the use of the occupant thermostat setpoint override model in conjunction with heat pump power modulation.

\subsection{Control task distribution in CityLearn v2} \label{sec:examples-control_task_distribution}
The work in \cite{simbuild2024_2283} showcases 17 different control tasks (\cref{tab:simbuild_24_simulation_matrix}) of differing complexity that are solved in CityLearn v2 using a subset of The CityLearn Challenge 2023 datasets. Complexity refers to (1) the number of controllable \glspl{der} present in a building including \gls{dhw} \gls{tes} (\verb|dhw_storage|), \gls{bess} (\verb|electrical_storage|) when paired with a \gls{pv} system and heat pump (\verb|cooling_device|), (2) the simplicity of the control algorithm i.e., explainable \gls{rbc} or adaptive but black-box \gls{rlc}, (3) the control objective (reward function for \gls{rlc}) designed for reduction in electricity consumption, energy cost, \gls{ghg} emission, discomfort, peak demand, or a combination of the aforementioned objectives, and (4) the size of the district i.e., number of buildings.

\begin{table*}
    \centering
    \footnotesize
    \caption{Summary of CityLearn v2 control tasks in \cite{simbuild2024_2283} describing the number of buildings in controlled district, \gls{kpi} to minimize, control agent and controlled \glspl{der}.}
    \label{tab:simbuild_24_simulation_matrix}
    \begin{tabularx}{\textwidth}{lrXlcccc}
        \toprule
        \textbf{ID} & \textbf{Buildings} & \textbf{KPI} & \textbf{Control} & \textbf{DHW TES} & \textbf{BESS} & \textbf{PV} & \textbf{Heat pump} \\
        \midrule
        \textit{x-b1\_b2-x-x} & 2 & - & Baseline & & & & \\
        \textit{x-b1\_b2-x-pv} & 2 & - & Baseline & & & \ding{51} & \\
        \textit{rbc-b1-c-dhw} & 1 & Cost & \gls{rbc} & \ding{51} & & & \\
        \textit{rbc-b1-e-dhw} & 1 & Emissions & \gls{rbc} & \ding{51} & & & \\
        \textit{rbc-b1-c-bess\_pv} & 1 & Cost & \gls{rbc} & & \ding{51} & \ding{51} & \\
        \textit{rbc-b1-e-bess\_pv} & 1 & Emissions & \gls{rbc} & & \ding{51} & \ding{51} & \\
        \textit{rbc-b1-c-dhw\_bess\_pv} & 1 & Cost & \gls{rbc} & \ding{51} & \ding{51} & \ding{51} & \\
        \textit{rbc-b1-e-dhw\_bess\_pv} & 1 & Emissions & \gls{rbc} & \ding{51} & \ding{51} & \ding{51} & \\
        \textit{rbc-b1\_b2-p-bess\_pv} & 2 & Peak& \gls{rbc} & & \ding{51} & \ding{51} & \\
        \textit{rlc-b1-c-dhw} & 1 & Cost & \gls{sac} & \ding{51} & & & \\
        \textit{rlc-b1-e-dhw} & 1 & Emissions & \gls{sac} & \ding{51} & & & \\
        \textit{rlc-b1-c-bess\_pv} & 1 & Cost & \gls{sac} & & \ding{51} & \ding{51} & \\
        \textit{rlc-b1-e-bess\_pv} & 1 & Emissions & \gls{sac} & & \ding{51} & \ding{51} & \\
        \textit{rlc-b1-c-dhw\_bess\_pv} & 1 & Cost & \gls{sac} & \ding{51} & \ding{51} & \ding{51} & \\
        \textit{rlc-b1-e-dhw\_bess\_pv} & 1 & Emissions & \gls{sac} & \ding{51} & \ding{51} & \ding{51} & \\
        \textit{rlc-b1-d\_o-hp} & 1 & Discomfort \& electricity consumption & \gls{sac} & & & & \ding{51} \\
        \textit{rlc-b1\_b2-p-bess\_pv} & 2 & Peak & \gls{sac} & & \ding{51} & \ding{51} & \\
        \bottomrule
    \end{tabularx}
\end{table*}

\subsubsection{Cost or emission reduction}
\begin{figure}
    \centering
    \begin{subfigure}[]{\columnwidth}
        \centering
        \includegraphics[width=\columnwidth]{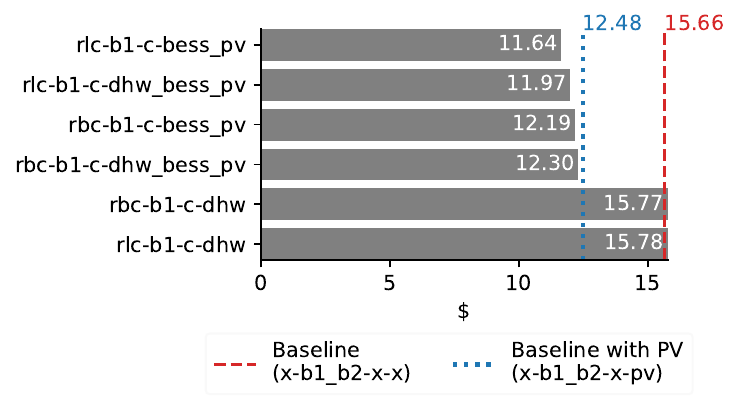}
        \caption{Cost (\$).}
        \label{simbuild_2024/cost_comparison}
    \end{subfigure}
    \begin{subfigure}[]{\columnwidth}
        \centering
        \includegraphics[width=\columnwidth]{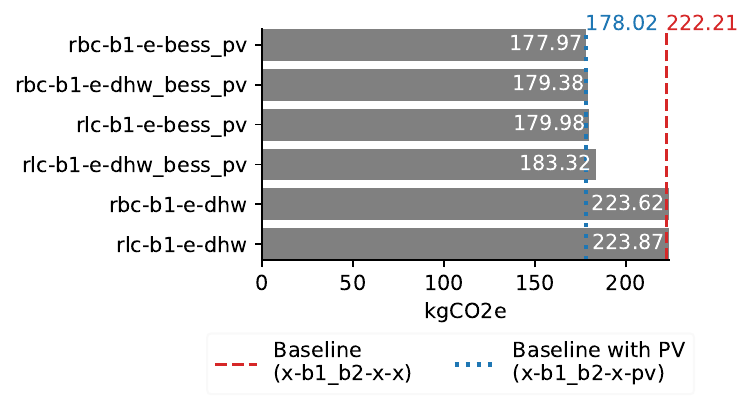}
        \caption{Emissions (kgCO\textsubscript{2}e).}
        \label{fig:simbuild_2024/emission_comparison}
    \end{subfigure}
    \caption{Cost (\$) and emissions (kgCO\textsubscript{2}e) results in \cite{simbuild2024_2283} from either \gls{rbc} or \gls{rlc} of \gls{dhw} \gls{tes}, \gls{bess}-\gls{pv} system or both when the control objective is to reduce cost or emission. The dashed red line shows the cost for a baseline scenario and the dotted blue line shows the cost for a baseline scenario but with solar generation to augment electricity from the grid.}
    \label{fig:simbuild_2024/cost_and_emission_comparison}
\end{figure}

\Cref{fig:simbuild_2024/cost_and_emission_comparison} shows cost (\$) and emissions (kgCO\textsubscript{2}e) in a building from \gls{rbc} or \gls{rlc} of \gls{dhw} \gls{tes}, \gls{bess}-\gls{pv} system or both when the control objective is energy cost (\cref{simbuild_2024/cost_comparison}) or emission (\cref{fig:simbuild_2024/emission_comparison}) reduction. The results are compared to baseline configurations without \gls{pv} (\textit{x-b1\_b2-x-x}) and with \gls{pv} (\textit{x-b1\_b2-x-pv}) in a 17-day test period. The \gls{pv} system advantage in terms of cost and emissions reduction is approximately 20.0\%, and is reflective of the 21.5\% \gls{zne} that the \gls{pv} system was sized for. The results show that \gls{rlc} provides further cost reduction of up to 6.7\% compared to \gls{rbc} while, neither algorithm provides substantial reduction in emissions irrespective of what \glspl{der} are available. These result from using emissions as control signal highlighted the importance of simultaneously decarbonizing the supply-side as demand-side end uses are electrified or risk an adverse effect of increased emissions. The variance in carbon emissions were less than 1.0\% and  renewable energy sources in the \gls{ercot} grid for the time period was estimated at only 20.1\% with 5.5\% standard deviation \cite{electric_reliability_council_of_texas_fuel_2021} making it a weak control signal.

\begin{figure*}
    \centering
    \includegraphics[width=\textwidth]{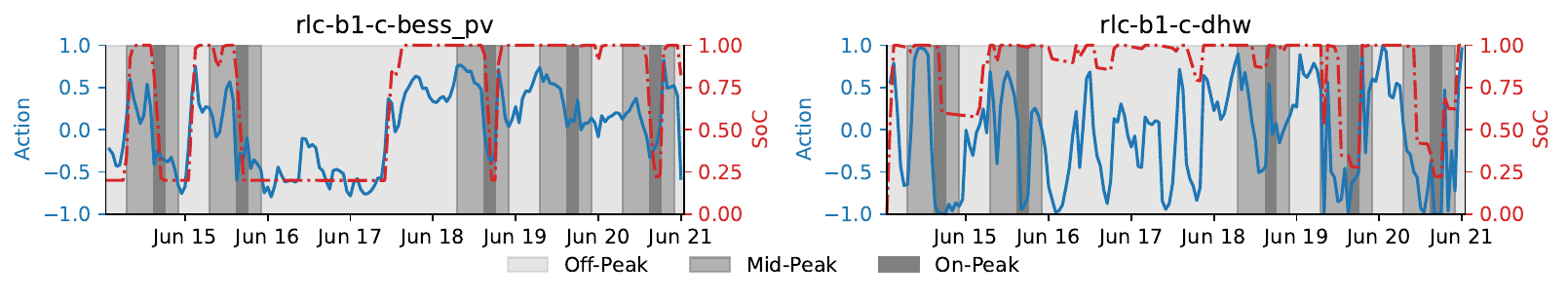}
    \caption{\Gls{rlc} action for \gls{bess} (left) and \gls{dhw} \gls{tes} (right), and consequent \gls{soc} trend in the initial seven days of the two-week evaluation period when the control objective is to minimize cost in \cite{simbuild2024_2283}.}
    \label{simbuild_2024/cost_soc_action_time_series}
\end{figure*}

\Cref{simbuild_2024/cost_soc_action_time_series} shows the \gls{rlc} actions in the initial seven days of the test period for the \textit{rlc-b1-c-dhw} and \textit{rlc-b1-c-dhw\_bess\_pv} configurations that have a cost reduction objective. The trend in the \gls{ess} \gls{soc} as a consequence of the actions are shown as well as shaded regions to indicate the \gls{tou} pricing. The \gls{soc} trend shows that the \gls{rlc} agent learns to charge the \glspl{ess} during the off-peak period while discharging mainly during the on-peak period. Compared to the \gls{bess}, the \gls{dhw} \gls{tes} rarely discharges its full capacity. This observation is attributed to the infrequent occurrence of \gls{dhw} loads where only 7\% of the total building load is attributed to \gls{dhw} load, as there is no \gls{dhw} load during 330/407 control time steps.

\subsubsection{Peak reduction}
\begin{figure}
    \centering
    \includegraphics[width=\columnwidth]{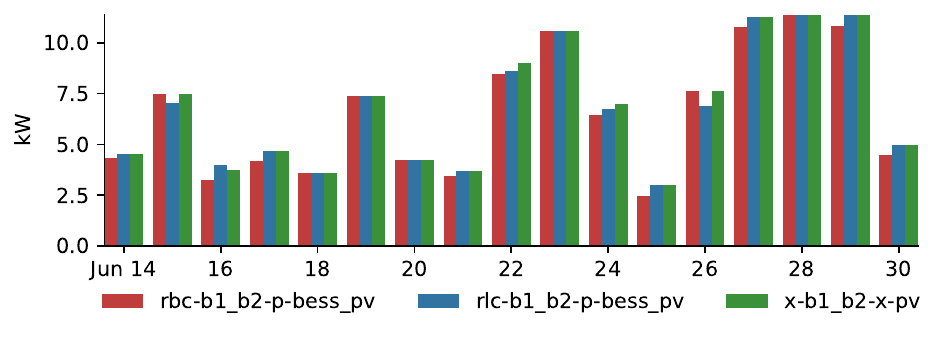}
    \caption{District-level daily peak load for two-building district where each building has \gls{bess}-\gls{pv} system in \cite{simbuild2024_2283}.}
    \label{simbuild_2024/daily_peak}
\end{figure}

The district-level daily peak load is shown in \cref{simbuild_2024/daily_peak} for the baseline with \gls{pv} generation (\textit{x-b1\_b2-x-pv}) and the two configurations with peak reduction objective but differing control algorithms. The \gls{rbc} in \textit{rbc-b1\_b2-p-bess\_pv} that has been fine-tuned to target energy discharge during peak periods provides a 2.6\% advantage in average daily peak reduction over the \textit{rlc-b1\_b2-p-bess\_pv} alternative that uses electricity consumption as a reward signal.

\subsubsection{Discomfort and electricity consumption reduction}
In the \textit{rlc-b1-d\_o-hp} simulation that modulates the \verb|cooling_device| heat pump power, a modified version of the comfort reward (\cref{eqn:reward_function-comfort}) is used where only cooling mode is considered. Exponents $a=1$ and $b=1$, and a penalty coefficient, $m$, is multiplied with the reward when there is over-cooling i.e., $T_t^{\textrm{in}} < T_t^{\textrm{spt}}$. Thus, comfort and energy efficiency can be learned by way of this penalty in the reward function for increased electricity consumption when over-cooling. \Cref{tab:simbuild_2024/discomfort_consumption_summary} shows the effects on discomfort and electricity consumption from tuning $m$. The percent changes in \glspl{kpi} by setting $m=3$, $m=6$, and $m=12$ are compared to the baseline case, $m=1$ (no penalty for over-cooling). By setting $m=3$, consumption and discomfort are reduced by 3.0\% by 50.0\% decrease in over-cooling. In contrast, $m=3$ increases under-cooling by only 20.0\%. Larger increments in $m$ show an increase in discomfort of almost 100.0\% from under-cooling compared to the baseline but up to 12.0\% decrease in consumption. However, the average temperature difference between the indoor dry-bulb temperature and setpoint for the selected $m$ values, irrespective of over-cooling or under-cooling, is less than 1.5C.

\begin{table*}
    \centering
    \begin{tabular}{rrrrr}
    \hline
         & \bf Discomfort & \bf Over-cool & \bf Under-cool & \bf Electricity consumption \\
          $\bm m$ & \bf ($^\circ$Ch) & \bf ($^\circ$C) & \bf ($^\circ$C) & \bf (kWh) \\
         \hline
        1 & 201.8 & $0.4\pm0.5$ & $0.5\pm0.5$ & 384.8 \\
        3 & 195.7 (\textcolor{blue}{-3.0\%}) & $0.2\pm0.2$ (\textcolor{blue}{-50.0\%}) & $0.6\pm0.5$ (\textcolor{red}{20.0\%}) & 373.0 (\textcolor{blue}{-3.0\%}) \\
        6 & 315.9 (\textcolor{red}{56.5\%}) & $0.3\pm0.2$ (\textcolor{blue}{-25.0\%}) & $0.9\pm0.5$ (\textcolor{red}{80.0\%}) & 352.5 (\textcolor{blue}{-8.3\%}) \\
        12 & 399.6 (\textcolor{red}{98.0\%}) & $0.1\pm0.4$ (\textcolor{blue}{-75.0\%}) & $1.0\pm0.6$ (\textcolor{red}{100.0\%}) & 338.1 (\textcolor{blue}{-12.0\%}) \\
        \hline
    \end{tabular}
    \caption{Effect of coefficient, $m$ on discomfort, average discomfort when over-cooling and under-cooling , and electricity consumption objectives when used in heat pump control simulation, \textit{rlc-b1-d\_o-hp} (\cite{simbuild2024_2283}). The percent change in \gls{kpi}, by varying $m$ is compared to the baseline case, $m=1$, where there is no penalty for increased consumption from over-cooling. Improvement in a \gls{kpi} is highlighted in \textcolor{blue}{blue} while deterioration is highlighted in \textcolor{red}{red}.}
    \label{tab:simbuild_2024/discomfort_consumption_summary}
\end{table*}

\subsection{Energy storage system control in representative single-family neighborhoods} \label{sec:examples-ess_control_in_neighborhoods}
In \cite{bs2023_1404}, the framework for designing virtual grid-interactive communities is applied to create three representative single-family neighborhoods in Alameda County, California (CA); Chittenden County, Vermont (VT); and Travis County, Texas (TX). For each location, six clusters of building metadata were found from which a frequency-based sampling is used to select up to 100 buildings. The CA, VT, and TX neighborhoods have 73, 43 and, 100 buildings respectively and each building is equipped with \gls{bess}-\gls{pv} system and \gls{dhw} \gls{tes} in an independent multi-agent \gls{rlc} architecture. The solar penalty reward, \cref{eqn:reward_function-solar_penalty}, that encourages net-zero energy use by penalizing load satisfaction from the grid when there is stored energy in \glspl{ess} as well as penalizing net export when \glspl{ess} are not fully charged is used to train the independent agents.

\begin{figure}
    \centering
    \includegraphics[width=\columnwidth]{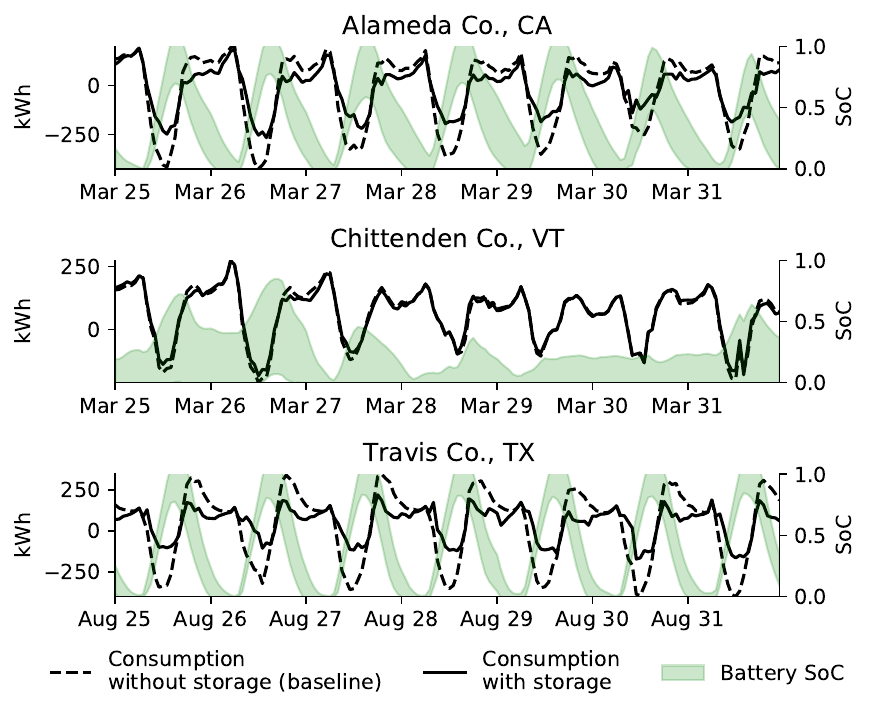}
    \caption{Results from \cite{bs2023_1404} comparing district-level net electricity consumption for scenarios with and without storage. The shaded regions show the average \gls{soc}\textsuperscript{BESS} $\pm$ one standard deviation.}
    \label{fig:bs_2023/neighborhood_electricity_consumption_profile_snapshot}
\end{figure}

\Cref{fig:bs_2023/neighborhood_electricity_consumption_profile_snapshot} compares the district-level net electricity consumption for scenarios with and without storage for a winter week in CA and VT neighborhoods, and a summer week in the TX neighborhood. The shaded regions show the average \gls{soc}\textsuperscript{BESS} $\pm$ one standard deviation. In CA and TX, the control agents learn to take advantage of the daytime solar generation to charge the \glspl{bess} and release the stored energy in the evening thus, reducing the daily peaks by 8.0\% and 42.0\% on average in either neighborhood. The VT profiles for the cases with and without storage are similar as the agents do not learn the load shifting task as in the case of the other two neighborhoods. The \gls{soc}\textsuperscript{BESS} distribution in VT shows irregular charge-discharge cycles and underutilized battery capacity for load shifting. This disparity in the VT results indicates the need for location-specific control strategies.

\subsection{Vehicle-to-grid control} \label{sec:examples-vehicle_to_grid}
The work in \cite{fonseca_evlearn_2024} demonstrates a \gls{v2g} scenario in CityLearn v2 for a nine-building residential and commercial building district including a medium-sized office, fast-food restaurant, standalone retail store, strip mall, and five medium-scale multifamily residences. The control signal is the real-time electricity pricing from the Iberian wholesale energy market (OMIE) in Portugal. The control algorithm, EnergAIze \cite{fonseca2024energaize} is a decentralized multi-agent reinforcement learning and is benchmarked against a baseline where there are no \gls{ev} loads. A synthetic 12-\gls{ev} schedule is used to train the \gls{rl} agent to prioritize. 

\begin{figure}
    \centering
    \includegraphics[width=\linewidth]{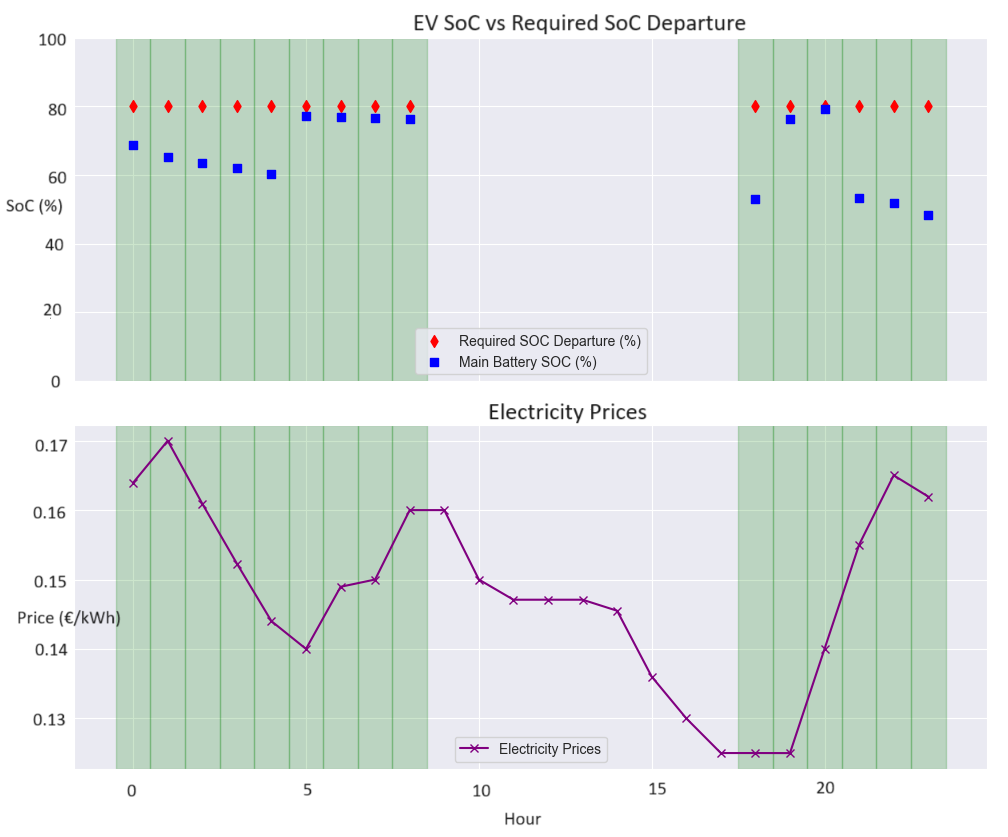}
    \caption{Controlled \gls{ev} \gls{soc} in \gls{v2g} control scenario (top) to reduce energy cost for a real-time pricing scheme (bottom). The green regiod is when the \gls{ev} is connected to a charger.}
    \label{fig:citylearn_control_example}
\end{figure}

\Cref{fig:citylearn_control_example} shows an \gls{ev}'s  controlled \gls{soc} with respect to expected departure \gls{soc} and real-time pricing for a day, where the green region is when the \gls{ev} is connected to a charger. At 8 am, the \gls{ev} leaves with an \gls{soc} close to what is expected at departure. On arrival (6 pm onwards), the \gls{rl} agent learns to charge when pricing is low and to discharge to the building when pricing is high.

\subsection{Occupant comfort feedback during automated demand response} \label{sec:examples-occ_in_control}
A case study is presented in \cite{kaspar_occ_2024}, where the heat pumps of ten single-family homes are controlled for \gls{dr} during a three-month winter heating period in Montreal, Quebec. The study included two occupant types, Average and Tolerant, based on different thermostat temperature setpoint preferences, and the integration of the occupants into the control scheme was modeled using three \glspl{lod}. \Gls{lod} 1 represents the baseline scenario where fixed setpoint schedules are used for each occupant type with no control of the HVAC system. \gls{lod} 2 also uses a fixed setpoint schedule but assumes occupant thermal comfort within a band of $2^\circ\mathrm{C}$ from the setpoint schedule following the methodology used in \cite{pinto_data-driven_2021}, thus allowing for flexibility of the HVAC system. In the most detailed \gls{lod}, \gls{lod} 3 incorporates occupant-thermostat override models and \gls{dr} events. The setpoint is reduced by $1.1^\circ\mathrm{C}$ ($2^\circ\mathrm{F}$) during the three or four-hour \gls{dr} events and the occupants can adjust the setpoint at each hourly time step due to thermal discomfort, which is dictated by the occupant-thermostat override models.

\begin{figure}
    \centering
    \includegraphics[width=\columnwidth]{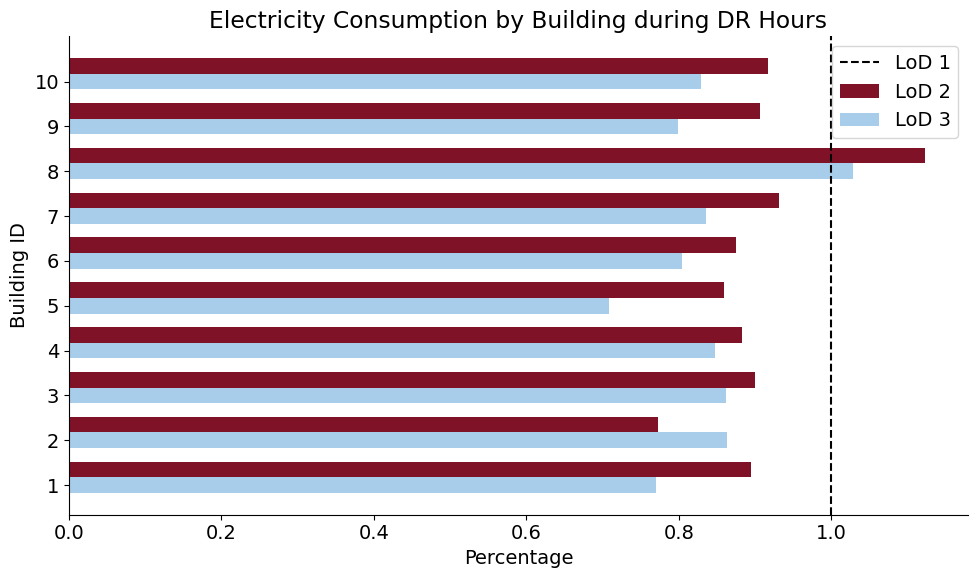}
    \caption{Total electricity consumption during \gls{dr} hours for \glspl{lod} 2 and 3 with respect to \gls{lod} 1.}
    \label{fig:electricity_consumption_dr}
\end{figure}

The \gls{eulp} dataset provides input data for building energy models in CityLearn, and the objective is to determine the flexibility services provided through the automated control of residential heating systems
in a ten-home neighborhood through setpoint adjustments during \gls{dr} events. In \gls{lod} 1 there is no controlled resource, while for \gls{lod} 2 and \gls{lod} 3 the \gls{sac} \gls{rl} algorithm \cite{haarnoja_soft_2018} is used to control the heat pump power to deliver adequate heating energy to each building to maintain indoor temperature in the comfort range for the occupant while providing flexibility. The control is evaluated using grid-level \glspl{kpi} including cost, total electricity consumption, and average daily peak, while also quantifying occupant discomfort through the number of thermostat overrides made. \Cref{fig:electricity_consumption_dr} shows the electricity consumption by household during the \gls{dr} events, during which \gls{lod} 3 implemented thermostat setbacks to specifically target electricity reductions during these hours while \gls{lod} 1 and \gls{lod} 2 use the normal setpoint schedules. The results show that during the \gls{dr} events, the electricity consumption was reduced by approximately 17\% on average during these peak hours for \gls{lod} 3, even with some occupants overriding the setpoint changes during the \gls{dr} events.

\begin{figure}
    \centering
    \includegraphics[width=\columnwidth]{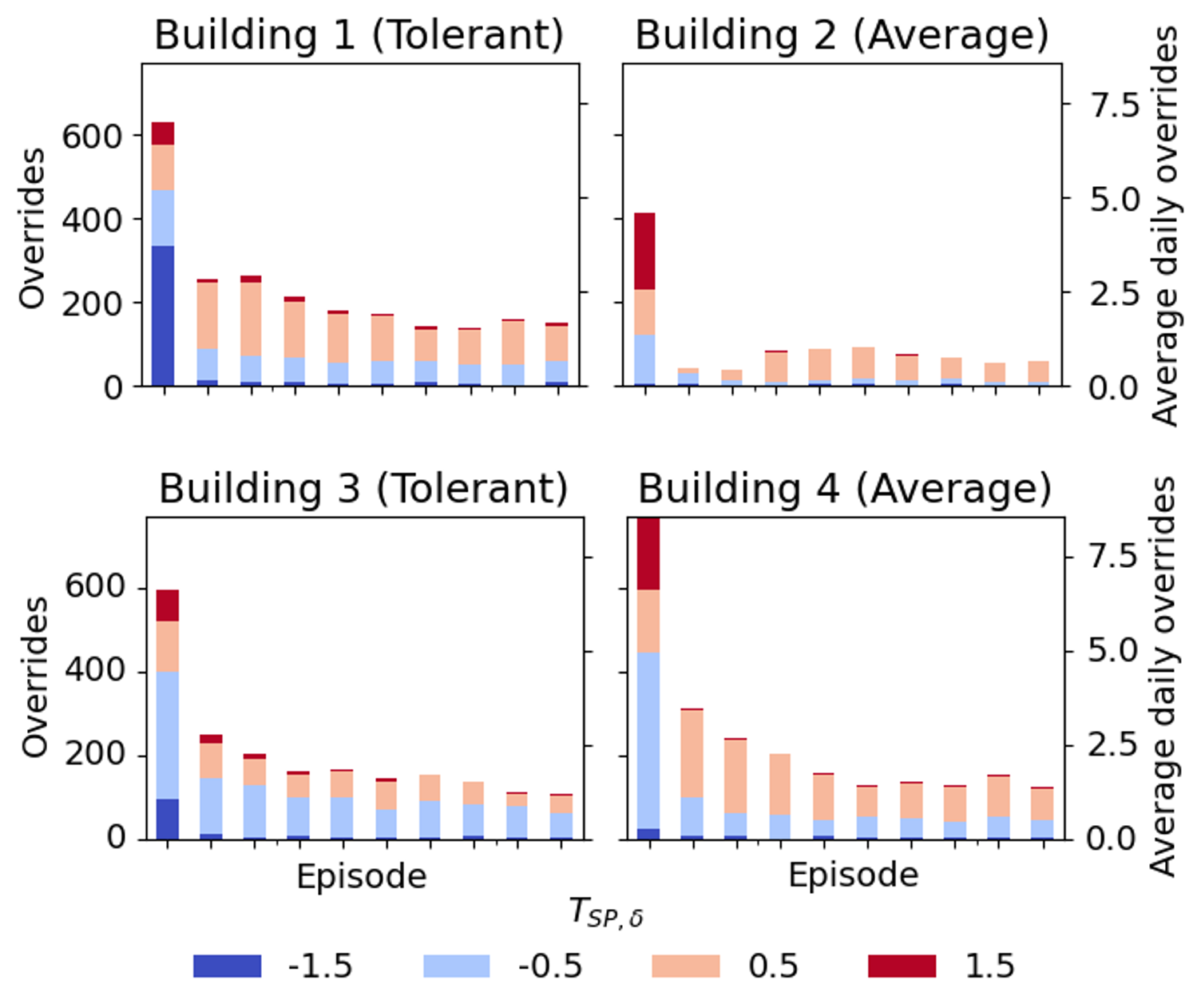}
    \caption{Number of occupant overrides and magnitude of the override for the 10 training episodes (adapted from \cite{kaspar_occ_2024}).}
    \label{fig:training_episodes_overrides}
\end{figure}

In addition to energy efficiency metrics, the agents are able to also reduce the number of thermostat overrides over time. \Cref{fig:training_episodes_overrides} shows the number of thermostat overrides throughout the ten training episodes for Buildings 1 and 3 with Tolerant occupants and Buildings 2 and 4 with Average occupants, where the darker red and blue represent larger changes to the setpoint and the light red and blue represent smaller changes to the setpoint. The figure shows that the agents were thus able to learn the best actions while considering the unique preferences and override behaviors of the two occupant types, and by using a multi-agent approach, the agents are able to learn optimal decision-making that achieves both occupant thermal comfort and \gls{dr} goals.

\section{Discussion} \label{sec:discussion}
The heat pump power control example that balances comfort and electricity use (see \cref{tab:simbuild_2024/discomfort_consumption_summary}) shows that one \gls{kpi} may be improved at the expense of the other. Thus, it is important to have a comprehensive quantification of flexibility caused by a control solution, including the adverse effects it may have. Methods for energy flexibility quantification is an open research question \cite{li_ten_2022}. The work by \citeauthor{li_data-driven_2023} has focused on bridging this research gap by identifying 48 data-driven energy flexibility \glspl{kpi} in the literature \cite{li_data-driven_2023}. These \glspl{kpi} are further classified as baseline-free and baseline-required where the latter are less suited for real world deployment because they require identical boundary conditions as the control event scenario for comparison. Ongoing work in \cite{bs2023_1305} is to provide these \glspl{kpi} in a Python package for integration into simulation environments such as CityLearn. Thus, a future CityLearn release will integrate this package to provide users with a plethora of energy flexibility \glspl{kpi} that better inform their control design.

In the same vein of realistic control deployment, the observations listed in \cref{tab:citylearn_observation_space}, though exhaustive and indeed observable, may be cost prohibitive to sense or calculate in practice \cite{blum_building_2021}. An example of such costly observation is solar irradiance where few sources provide real-time irradiance data and in most cases are behind a paywall. Similarly, the real time cooling, heating and \gls{dhw} loads in a building without an accompanying digital twin are challenging to measure compared to the electricity consumption that these loads depend on. Also, \gls{ev} departure time is not always accurately known apriori without deliberate insight from the occupant. Yet, the simulation and real-world environments should have similar control spaces for reproducibility of simulation results in practice. Thus, the control designer must consider the practical implications of their observation space on real world adaptation.

The example in \cref{sec:examples-ess_control_in_neighborhoods} assessed energy flexible neighborhoods in only three climate zones with a limited control period of three months in the winter or summer. While the work demonstrates the efficacy of the framework, it is limited in providing conclusive results on energy flexibility of the U.S. residential building stock for the current and future climatic conditions. Also, the considered controlled \glspl{der} are limited to only \gls{ess} for load shifting without consideration of flexibility provided by other \glspl{der} e.g. heat pumps and \glspl{ev} nor consideration for control resilience and occupant model in the control loop. Future work will conduct a comprehensive analysis of various \gls{der} availability configurations in representative neighborhoods of at least, the 16 climate zones classifications by ASHRAE that differ in temperature and humidity conditions, to provide a holistic view of the energy flexibility potential of the residential building stock.

The \gls{v2g} control example shows the potential benefits of increased \gls{ev} adoption when properly managed. The observation across the different chargers in the district for the entire four-year simulation period is an 11\% reduction in energy cost and 21\% reduction in average daily peak (see \cite{fonseca_evlearn_2024}). With policies aimed at replacing internal combustion engine vehicle fleets with \glspl{ev}, the \gls{ev} share of controllable \glspl{der} will increase. There has been a rise of \gls{ev} sales from 4\% of total worldwide car sales in 2020 to over 14\% in 2022 \cite{iea2023}. These sales are incentivized by tax credits \cite{jenn_effectiveness_2018}, access to carpool lanes, reduced car registration fees, and subsidies for constructing residential charging stations \cite{ghose_studying_2023}.

The work on integrating the occupant thermal comfort feedback presented in \cref{sec:examples-occ_in_control}, it is important to note that all occupant modeling was done using ecobee's Donate Your Data dataset \cite{luo_ecobee_2022} for homes in Quebec, Canada during the winter heating season. As such, future work would be required to determine the occupant models' applicability in other climates, and additional location-specific override models could be developed using the methodology presented in the study. However, one of the key contributions of this work is showing the impact that the occupant behavior \gls{lod} has on the load shifting potential and occupant thermal comfort. We provide several examples via the \glspl{lod} of how one may incorporate occupants in the control scheme that reflect a more realistic scenario of a \gls{dr} program in which occupants are allowed to opt out of thermostat setbacks during the \gls{dr} events.

While existing literature has demonstrated control scalability in CityLearn \cite{almilaify_scalex_2023}, a challenge of large scale analysis especially using \gls{rlc} is that it may be computationally and data cost prohibitive to train a unique control policy for each building. Transfer learning can alleviate these costs by leveraging the trained policy of a source building to initialize those of the remaining target buildings with outcomes that are better or comparable to baseline and reference control policies \cite{nweye_martini_2022,coraci_online_2023}. On the other hand, with the diversity of control tasks i.e., controllable \glspl{der}, meta-learning can provide a generalized algorithm-agnostic control policy that can quickly adapt to any single control task \cite{beck_survey_2023}. Yet, there are few applications of meta-learning in building control and demand response \cite{mason_review_2019}. A meta-policy will allow for decentralized heterogeneous control that depicts real world large scale deployment of advanced controllers where different buildings, and in special cases, different \glspl{der}, use unique control algorithms, observation and action spaces \cite{wu_heterogeneous_2023}.

We note that the heat pump model in CityLearn assumes an ideal refrigeration cycle and the \gls{cop} is temperature-driven alone. Also, the control signal for the heat pump at this time does not allow for thermostatic control of the heat pump through setpoints which could limit its suitability for true demand response applications. Thus, CityLearn is best applied to simple and quick comparative analysis between control algorithms with minimal design cost. Whereas, an environment that utilizes high-fidelity energy models e.g., OCHRE \cite{blonsky_ochre_2021} is better suited for analyses that are purposed for scrutiny of absolute energy and cost values. However, the challenge with using these high-fidelity environments is their computational overhead cost.

One limitation of CityLearn is that there are no district energy system models such as centralized \gls{pv} system, \gls{ess}, geothermal cooling or heating infrastructure. However, from a control perspective, the buildings can share information about their states to achieve district-level energy flexibility. Future work could consider the integration of such systems in the environment with reference to simulation environments like CityBES \cite{hong_citybes_2016} and URBANopt \cite{el_kontar_urbanopt_2020}. Future development of CityLearn may also include adding multi-zone building thermal models that allow simultaneous cooling and heating as well as consideration of indoor humidity loads, and adding indoor air quality models and consideration of \gls{ieq} metrics in the \glspl{kpi}. Field validation of CityLearn in real building districts is also needed to demonstrate its usability and performance, as well as to inform the development of best practice guides.

Despite its potential, there are still open questions regarding the plug-and-play capabilities, performance, safety of operation, and learning speed of \gls{rlc} \cite{dulac-arnold_challenges_2021}. The CityLearn Challenge is an avenue for crowd-sourcing solutions to building control problems, using any control algorithm of choice. It attracts a multidisciplinary participation including researchers, industry experts, sustainability enthusiasts and \gls{ai} hobbyists with a focus on \gls{der} control algorithm benchmarking for demand response and energy flexibility settings. Previous editions of The CityLearn Challenge have investigated control policy transferrability across different climate zones \cite{vazquez-canteliCityLearnChallenge2020-2}, a realistic implementation of model-free \gls{rlc} in buildings where training and evaluation were done on a single four-year episode \cite{nagy_citylearn_2021}, and \gls{bess} control using a real-world \gls{zne} community dataset for cost and emissions optimization \cite{nweye_citylearn_2022}. Other competitions in the energy domain focus on building load predictions \cite{miller_ashrae_2020}, grid power flow optimization \cite{holzer_grid_2021} and pathways to building electrification and decarbonization \cite{nyserda_nyserda_2022}. Thus, we envision opportunities for collaboration amongst these different competitions where their unique objectives and experience, can provide holistic solutions to the problems in the building energy and control communities.
\section{Conclusion} \label{sec:conclusion}
In this paper, we introduce CityLearn v2, an environment based on the Farama Foundation Gymnasium interface, that leverages the End-Use Load Profiles for the U.S. Building Stock dataset to create virtual grid-interactive communities for benchmarking resilient multi- agent, \acrlong{der} and objective control with dynamic occupant feedback. CityLearn allows for custom building districts as users can select which buildings to include, \acrlongpl{der} to control, and observations to sense. Users can also design custom reward functions and choose between three control configurations: central single-agent, independent multi-agent and coordinated multi-agent with provision of in-built control algorithms and support for third-party control libraries. These modelling options improve the level of realism in CityLearn, which is an important guideline when designing these kinds of environments \cite{wolfle_guide_2020}.

\citeauthor{rolnick_tackling_2022} investigate how the machine learning community can help tackle climate change \cite{rolnick_tackling_2022}. The authors identify electrification of urban energy systems as a field where machine learning can provide high leverage. The review of use cases of machine learning in the building's life cycle by \citeauthor{hong_ml} indicates promising potential of machine learning based smart controls for grid-interactive efficient buildings \cite{hong_ml}. CityLearn is aimed precisely at this objective, and a standardized environment, distributed as educational material \cite{nweye2023citylearn}, will allow more researchers from the computer science and machine learning communities to participate in the search for solutions to tackle climate change. Also, by simplifying the \acrlong{bes} and providing different environment datasets, CityLearn can help building energy engineers and control engineers to focus on control algorithm development with less time spent setting up the environment.

Lastly, the application examples presented in this work are by no means an exhaustive demonstration of CityLearn's functionalities. Instead, we envision that our contribution provides environment and control examples as well as their implementation source code for actors in the building control community who are interested in using CityLearn to solve their \acrlong{der} control problems.



\section{Declaration of competing interest}
The authors certify that they have no affiliations with or involvement in any organization or entity with any financial interest, or non-financial interest in the subject matter or materials discussed in this manuscript.

\bibliographystyle{cas-model2-names}

\bibliography{references-kingsley,references-other}

\clearpage
\appendix
\renewcommand\thefigure{\thesection.\arabic{figure}}
\renewcommand\thetable{\thesection.\arabic{table}}
\renewcommand\theequation{\thesection.\arabic{equation}}

\section{Appendix}\label{sec:appendix}
\setcounter{figure}{0}
\setcounter{table}{0}
\setcounter{equation}{0}

\subsection{Distributed energy resources} \label{sec:environment-distributed_energy_resources}
This section describes the energy models of the \glspl{der} in CityLearn to include the heat pump \cref{sec:environment-heat_pump}, electric heater \cref{sec:environment-electric_heater}, \gls{tes} (\cref{sec:environment-thermal_energy_storage}), \gls{bess} (\cref{sec:environment-battery_model}), \gls{ev} (\cref{sec:environment-electric_vehicle}), and \gls{pv} system (\cref{sec:environment-pv}). All \glspl{der} but the \gls{pv} system are controllable.

\subsubsection{Heat pump} \label{sec:environment-heat_pump}
CityLearn uses an air-to-water heat pump model that is based on an ideal refrigeration cycle. Thus, the \gls{cop} (\cref{eqn:heat_pump-cooling_cop,eqn:heat_pump-heating_cop}) at any given time step, $t$, is a function of only outdoor air temperature, $T^{\textrm{out}}$, and target supply temperatures for cooling and heating modes ($T^{\textrm{cooling, supply}}$, $T^{\textrm{heating, supply}}$), where the supply temperature is constant at all time steps. The heat pump can be used in either control or ideal mode. In the control mode, the electricity consumption of the heat pump, $E_t^{\textrm{HP, control}}$, is a function of the control action $a_{t - 1}^{\textrm{HP}}$ where $a \in [0, 1]$ denotes the proportion of heat pump's nominal power, $P^{\textrm{HP, nominal}}$, made available (\cref{eqn:heat_pump-electricity_consumption-control}). The supplied cooling or heating energy for the controlled power is then defined by \cref{eqn:heat_pump-output_energy} as the product of the \gls{cop}, electricity consumption and technical efficiency, $\eta^{\textrm{HP, technical}}$.

\begin{equation}
    \textrm{COP}_t^{\textrm{cooling}} = \frac{T^{\textrm{cooling, supply}}}{T_t^{\textrm{out}} - T^{\textrm{cooling, supply}}}
    \label{eqn:heat_pump-cooling_cop}
\end{equation}

\begin{equation}
    \textrm{COP}_t^{\textrm{heating}} = \frac{T^{\textrm{heating, supply}}}{T^{\textrm{heating, supply}} - T_t^{\textrm{out}}}
    \label{eqn:heat_pump-heating_cop}
\end{equation}

\begin{equation}
    E_t^{\textrm{HP, control}} = a_{t - 1}^{\textrm{HP}} \times P^{\textrm{HP, nominal}}
    \label{eqn:heat_pump-electricity_consumption-control}
\end{equation}

\begin{equation}
    Q_t^{\textrm{HP, control}} = \eta^{\textrm{HP, technical}} \times \textrm{COP}_t^{\textrm{cooling | heating}} \times E_t^{\textrm{HP, control}}
    \label{eqn:heat_pump-output_energy}
\end{equation}

In the ideal mode, the supplied cooling or heating energy is known apriori from \gls{beps} software e.g. EnergyPlus \cite{crawley_energyplus_2001}, or real world measurements, and only the electricity consumption for the supplied energy needs to be calculated as defined in \cref{eqn:heat_pump-electricity_consumption-ideal}.

\begin{equation}
    E_t^{\textrm{HP, ideal}} = \frac{Q_t^{\textrm{HP, ideal}}}{\eta^{\textrm{technical}} \times \textrm{COP}_t^{\textrm{cooling | heating}}}
    \label{eqn:heat_pump-electricity_consumption-ideal}
\end{equation}

\subsubsection{Electric heater} \label{sec:environment-electric_heater}
The electric heater is defined by \cref{eqn:electric_heater-electricity_consumption-control,eqn:electric_heater-output_energy,eqn:electric_heater-electricity_consumption-ideal} where like the heat pump, can be used in control or ideal mode. Similarly, in the control mode, the electricity consumption of the electric heater is a function of the control action $a_{t - 1}^{\textrm{EH}}$ where $a \in [0, 1]$ denotes the proportion of heater's nominal power, $P^{\textrm{EH, nominal}}$, provided for heating (\cref{eqn:electric_heater-electricity_consumption-control}). The supplied heating energy is the product of the heater's technical efficiency, $\eta^{\textrm{EH, technical}}$, and controlled electricity consumption, $E_t^{\textrm{EH, control}}$ (\cref{eqn:electric_heater-output_energy}). In the ideal case, $Q_t^{\textrm{EH, ideal}}$ is known and $E_t^{\textrm{EH, ideal}}$ evaluated as \cref{eqn:electric_heater-electricity_consumption-ideal}.

\begin{equation}
    E_t^{\textrm{EH, control}} = a_{t - 1}^{\textrm{EH}} \times P^{\textrm{EH, nominal}}
    \label{eqn:electric_heater-electricity_consumption-control}
\end{equation}

\begin{equation}
    Q_t^{\textrm{EH, control}} = \eta^{\textrm{EH, technical}} \times E_t^{\textrm{EH, control}}
    \label{eqn:electric_heater-output_energy}
\end{equation}

\begin{equation}
    E_t^{\textrm{EH, ideal}} = \frac{Q_t^{\textrm{EH, ideal}}}{\eta^{\textrm{EH, technical}}}
    \label{eqn:electric_heater-electricity_consumption-ideal}
\end{equation}

\subsubsection{Thermal energy storage} \label{sec:environment-thermal_energy_storage}
Thermal loads are shifted by storing (charging) and releasing (discharging) chilled or hot water energy in a \gls{tes}. The stored energy at any time step, $Q_t^{tes}$, is a piecewise function (\cref{eqn:thermal_energy_storage-energy}) driven by the control action $a_{t - 1}^{\textrm{TES}}$, where $a \in [-1, 1]$ prescribes the proportion of the \gls{tes} capacity, $C^{\textrm{TES}}$, to be charged ($a > 0$) or discharged ($a < 0$). The energy stored after charging is defined by \cref{eqn:thermal_energy_storage-energy-charging} as the minimum of the capacity and an expected energy. The expected energy is the sum of the initial energy after losses, $Q_{t - 1}^{\textrm{TES}} \times (1 - \theta^{\textrm{TES}})$, where $\theta^{\textrm{TES}}$ is the thermal loss coefficient, and the product of the action, capacity and round-trip efficiency, $\eta^{\textrm{TES, round-trip}}$. The round-trip efficiency defined in \cref{eqn:thermal_energy_storage-round_trip_efficiency} is the square root of the technical efficiency, $\eta^{\textrm{TES, technical}}$.

\begin{equation}
    Q_t^{\textrm{TES}} = \begin{cases}
        \cref{eqn:thermal_energy_storage-energy-charging} & \text{ if } a_{t - 1}^{\textrm{TES}} > 0.0 \\
        \cref{eqn:thermal_energy_storage-energy-discharging} & \text{ if } a_{t - 1}^{\textrm{TES}} < 0.0 \\
        0.0 & \text{ otherwise}
    \end{cases}
    \label{eqn:thermal_energy_storage-energy}
\end{equation}

\begin{multline}
    Q_t^{\textrm{TES, +}} = \textrm{min}\Bigg(C^{\textrm{TES}}, Q_{t - 1}^{\textrm{TES}} \times \Big(1 - \theta^{\textrm{TES}}\Big) \\
        + a_{t - 1}^{\textrm{TES}} \times C^{\textrm{TES}} \times \eta^{\textrm{TES, round-trip}} \Bigg)
    \label{eqn:thermal_energy_storage-energy-charging}
\end{multline}

\begin{equation}
    \eta^{\textrm{TES, round-trip}} = \sqrt{\eta^{\textrm{TES, technical}}}
    \label{eqn:thermal_energy_storage-round_trip_efficiency}
\end{equation}

The stored energy after discharging the \gls{tes} is defined in \cref{eqn:thermal_energy_storage-energy-discharging} as the sum of the initial energy after losses and quotient of the energy equivalent of the control action ($a_{t - 1}^{\textrm{TES}} \times C^{\textrm{TES}}$) and the round-trip efficiency. A lower limit of 0 is imposed to satisfy the energy balance in the case of a completely discharged \gls{tes}.

\begin{multline}
    Q_t^{\textrm{TES, -}} = \textrm{max}\Bigg(0.0, Q_{t - 1}^{\textrm{TES}} \times \Big(1 - \theta^{\textrm{TES}}\Big) \\
        + a_{t - 1}^{\textrm{TES}} \times C^{\textrm{TES}} \div \eta^{\textrm{TES, round-trip}} \Bigg)
    \label{eqn:thermal_energy_storage-energy-discharging}
\end{multline}

Consequently, the \gls{soc} (\cref{eqn:thermal_energy_storage-state_of_charge}) $\in [0, 1]$ is the ratio of the stored energy calculated in \cref{eqn:thermal_energy_storage-energy} and the capacity.

\begin{equation}
    \textrm{SoC}_t^{\textrm{TES}} = \frac{Q_t^{\textrm{TES}}}{C^{\textrm{TES}}}
    \label{eqn:thermal_energy_storage-state_of_charge}
\end{equation}

Finally, the energy balance i.e., energy supplied to the \gls{tes} by an electric device e.g., heat pump or electric heater before adjustment for efficiency or energy supplied from the \gls{tes} to a building after adjustment for efficiency is defined by \cref{eqn:thermal_energy_storage-energy_balance}.

\begin{multline}
    Q_t^{\textrm{TES, balance}} = \Big(Q_t^{\textrm{TES}} - Q_{t - 1}^{\textrm{TES}} \times \Big(1 - \theta^{\textrm{TES}}\Big)\Big) \\
        \times \begin{cases}
            1 \div \eta^{\textrm{TES, round-trip}} & \text{ if } a_{t - 1} > 0.0 \\
            \eta^{\textrm{TES, round-trip}} & \text{ if } a_{t - 1} < 0.0 \\
            0.0 & \text{ otherwise}
                \end{cases}
    \label{eqn:thermal_energy_storage-energy_balance}
\end{multline}

\subsubsection{Battery energy storage system} \label{sec:environment-battery_model}
The \gls{bess} model builds up on the \gls{tes} model but has a time-dependent capacity, $C^{\textrm{BESS}}$, round-trip efficiency, $\eta^{\textrm{BESS, round-trip}}$, and maximum input as well as output power, $P^{\textrm{BESS, max}}$. The stored energy at any time step, $Q_t^{bess}$, is a piecewise function (\cref{eqn:battery-energy}) driven by the control action $a_{t - 1}^{\textrm{BESS}}$, where $a \in [-1, 1]$ prescribes the proportion of the maximum capacity (before degradation), $C_0^{\textrm{BESS}}$, to be charged ($a > 0$) or discharged ($a < 0$). The stored energy after charging is defined by \cref{eqn:battery-energy-charging} as the minimum of the degraded capacity, $C_t^{\textrm{BESS}}$ (\cref{eqn:battery-energy-capacity}), and an expected energy. The expected energy is the sum of the initial energy after losses, $Q_{t - 1}^{\textrm{BESS}} \times (1 - \theta^{\textrm{BESS}})$, and the energy to be added, where $\theta^{\textrm{BESS}}$ is the thermal loss coefficient. This added energy is the product of the power-dependent round-trip efficiency, $\eta^{\textrm{BESS, round-trip}}$, and the minimum of two quantities: the energy equivalent of the control action ($a_{t - 1}^{\textrm{BESS}} \times P^{\textrm{BESS, nominal}}$) and the \gls{soc}-dependent maximum input and output power, $P_t^{\textrm{BESS, max}}$.

\begin{equation}
    Q_t^{\textrm{BESS}} = \begin{cases}
        \cref{eqn:battery-energy-charging} & \text{ if } a_{t - 1}^{\textrm{BESS}} > 0.0 \\
        \cref{eqn:battery-energy-discharging} & \text{ if } a_{t - 1}^{\textrm{BESS}} < 0.0 \\
        0.0 & \text{ otherwise}
    \end{cases}
    \label{eqn:battery-energy}
\end{equation}

\begin{multline}
    Q_t^{\textrm{BESS, +}} = \textrm{min}\Bigg(C_t^{\textrm{BESS}}, Q_{t - 1}^{\textrm{BESS}} \times \Big(1 - \theta^{\textrm{BESS}}\Big) \\
        + \textrm{min}\Big(a_{t - 1}^{\textrm{BESS}} \times C_0^{\textrm{BESS}}, P_t^{\textrm{BESS}}\Big) \\
            \times \eta^{\textrm{BESS, round-trip}} \Bigg)
    \label{eqn:battery-energy-charging}
\end{multline}

The stored energy after discharging the \gls{bess} is defined by \cref{eqn:battery-energy-discharging} as the sum of the initial energy after losses and quotient of the energy equivalent of the control action (after adjustment for maximum output power, $P_t^{\textrm{BESS}}$) and the round-trip efficiency. The stored energy is limited to a \gls{dod} such that the \gls{bess} is never completely drained for $\textrm{DoD} > 0$. \Cref{eqn:battery-state_of_charge} is the \gls{bess} \gls{soc} as a function of the stored energy and $C_0^{\textrm{BESS}}$.

\begin{multline}
    Q_t^{\textrm{BESS, -}} = \\
        \textrm{min}\Bigg(C_0^{\textrm{BESS}} \times \textrm{DoD}^{\textrm{BESS}}, Q_{t - 1}^{\textrm{BESS}} \times \Big(1 - \theta^{\textrm{BESS}}\Big) \\
            + \textrm{min}\Big(a_{t - 1}^{\textrm{BESS}} \times C_0^{\textrm{BESS}}, -P_t^{\textrm{BESS}}\Big) \\
            \div \eta^{\textrm{BESS, round-trip}} \Bigg)
    \label{eqn:battery-energy-discharging}
\end{multline}

\begin{equation}
    \textrm{SoC}_t^{\textrm{BESS}} = \frac{Q_t^{\textrm{BESS}}}{C_0^{\textrm{BESS}}}
    \label{eqn:battery-state_of_charge}
\end{equation}

The energy balance, \cref{eqn:battery-energy_balance}, is same as that of the \gls{tes} except that the round-trip efficiency is time-dependent. The energy balance is also the electricity consumed when charging or avoided consumption when discharging (\cref{eqn:battery-electricity_consumption}).

\begin{multline}
    Q_t^{\textrm{BESS, balance}} = \Big(Q_t^{\textrm{BESS}} - Q_{t - 1}^{\textrm{BESS}} \times \Big(1 - \theta^{\textrm{BESS}}\Big)\Big) \\
        \times \begin{cases}
            1 \div \eta_t^{\textrm{BESS, round-trip}} & \text{ if } a_{t - 1}^{\textrm{BESS}} > 0.0 \\
            \eta_t^{\textrm{BESS, round-trip}} & \text{ if } a_{t - 1}^{\textrm{BESS}} < 0.0 \\
            0.0 & \text{ otherwise}
                \end{cases}
    \label{eqn:battery-energy_balance}
\end{multline}

\begin{equation}
    E_t^{\textrm{BESS}} = Q_t^{\textrm{BESS, balance}}
    \label{eqn:battery-electricity_consumption}
\end{equation}

The time-dependency of the capacity is as a result of degradation from charging and discharging cycles. It is defined in \cref{eqn:battery-energy-capacity} as the difference between the capacity at the previous time step, and the the loss quantity. This quantity is a function of the capacity loss coefficient, $\phi$, maximum capacity, energy balance at the current time step, and capacity at the previous time step.

\begin{equation}
    C_t^{\textrm{BESS}} = C_{t - 1}^{\textrm{BESS}} - \frac{\phi \times C_0^{\textrm{BESS}} \times \Big\lvert Q_t^{\textrm{BESS, balance}} \Big\rvert}{2 \times C_{t - 1}^{\textrm{BESS}}}
    \label{eqn:battery-energy-capacity}
\end{equation}

\begin{figure}[]
    \centering
    \begin{subfigure}[t]{0.47\columnwidth}
        \centering
        \includegraphics[width=\columnwidth]{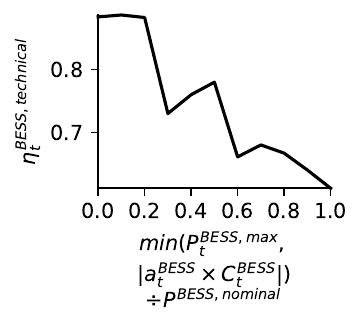}
        \caption{Technical efficiency as a function of controlled proportion of nominal power (\cref{eqn:battery-technical_efficiency}).}
        \label{fig:example_bess_power_efficiency_curve}
    \end{subfigure}
    \hspace{0.03\columnwidth}
    \begin{subfigure}[t]{0.47\columnwidth}
        \centering
        \includegraphics[width=\columnwidth]{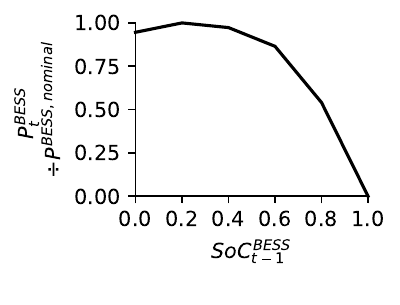}
        \caption{Proportion of nominal power used to define the maximum input and output power as a function of \gls{soc} (\cref{eqn:battery-maximum_power}).}
        \label{fig:example_bess_capacity_curve}
    \end{subfigure}
    \caption{Example curves defining \gls{bess} technical efficiency, $\eta_t^{\textrm{BESS, technical}}$, and maximum input and output power, $P_t^{\textrm{BESS, max}}$.}
    \label{fig:example_bess_curves}
\end{figure}

The \gls{bess} round-trip efficiency, $\eta_t^{\textrm{BESS, round-trip}}$, (\cref{eqn:battery-round_trip_efficiency}) is similar to that of the \gls{tes} but with the technical efficiency, $\eta_t^{\textrm{BESS, technical}}$, as a function of a power-efficiency curve (\cref{eqn:battery-technical_efficiency}), where for a given controlled proportion of the nominal power, $P^{\textrm{BESS, nominal}}$, there is a limitation on the battery's efficiency. An example of this function from \cite{lee_battery_2021} is shown in \cref{fig:example_bess_power_efficiency_curve} for a \gls{bess} subjected to continuous charge and discharge cycles.

\begin{equation}
    \eta_t^{\textrm{BESS, round-trip}} = \sqrt{\eta_t^{\textrm{BESS, technical}}}
    \label{eqn:battery-round_trip_efficiency}
\end{equation}

\begin{multline}
    \eta_t^{\textrm{BESS, technical}} = \\
        f\Big(a_t^{\textrm{BESS}}, C_0^{\textrm{BESS}}, P_t^{\textrm{BESS, max}}, P^{\textrm{BESS, nominal}}\Big)
    \label{eqn:battery-technical_efficiency}
\end{multline}

\Cref{eqn:battery-maximum_power} defines the maximum input and output power $P_t^{\textrm{BESS, max}}$ as the product of the nominal power, $P^{\textrm{BESS, nominal}}$, and \gls{soc}-power curve. The curve's dependent variable is the proportion of available nominal power, $P^{\textrm{BESS, nominal}}$, at any \gls{soc}, $\textrm{SoC}_{t-1}^{\textrm{BESS}}$ value.

\begin{equation}
    P_t^{\textrm{BESS, max}} = P^{\textrm{BESS, nominal}} \times f\Big(\textrm{SoC}_{t-1}^{\textrm{BESS}}\Big)
    \label{eqn:battery-maximum_power}
\end{equation}

\Cref{fig:example_bess_capacity_curve} shows an example of an \gls{soc}-dependent power curve from \cite{varesi_efficient_2016} where the \gls{bess} charges faster and close to the nominal power for lower \gls{soc} ($\textrm{SoC}_{t-1} \le 0.6$), while the charge rate decreases to zero at higher \gls{soc}. This is consistent with the charging behaviour of a \gls{bess} where it takes longer to charge per unit \gls{soc} after exceeding an \gls{soc} threshold of $\approx 0.8$ \cite{kostopoulos_real-world_2020}.

\subsubsection{Electric vehicle} \label{sec:environment-electric_vehicle}
The energy management of \glspl{ev} is designed to allow the simulation of three distinct modes: (1) \gls{v2g}; (2) \gls{g2v}; (3) no control (i.e., where \glspl{ev} act as a load without any possible control over their charging). For that purpose, the simulation has three fundamental parts: the Electric Vehicle Chargers (EVCs), which serve as a connection between a building and an \gls{ev}, the model of \gls{ev} itself, which acts as \gls{der}, and an input schedule, which dictates the plug in/out energy flexibility routine for each \gls{ev} and introduces formulation for energy flexibility and simulation modes. For a detailed explanation of the \gls{ev} modes and simulation please refer to ANONYMOUS.

\textbf{Electric vehicle charger:} \label{sec:environment-electric_vehicle_charger}
The EVC serves as an interface between the buildings in the simulation environment and the \glspl{ev}, and is the component where control is applied.

Just like a house, an office or any building can have multiple installed chargers in the real world, a single building in the simulation can have more than one charger simulated. Each charger is assigned a unique identifier $\textrm{EVC}_{b\_n\_p}$ (where EVC stands for Electric Vehicle Charger, b stands for the Building where the charger is inserted in the simulation, n for the charger Number within the building and p stands as the number of the Plug of that charger). This facilitates the appropriate linkage between the EVC and the EVs during the simulation. 

The model supports the specification of distinct power output levels for charging and discharging modes. When an EV is plugged in, the electricity consumption of the electric vehicle charger $E_t^{\textrm{EVC}_{b\_n\_p}}$ is a function of the control action $a_{t - 1}^{\textrm{EVC}_{b\_n\_p}}$ where $a \in [-1, 1]$ for V2G $a \in [0, 1]$ for G2V and No-Control modes. It denotes the proportion of charger's nominal power for charging $P^{\textrm{EVC, nominal, charging}}$ when $\text{if } a_{t - 1}^{\textrm{EVC}_{b\_n\_p}} \geq 0$ and the proportion of charger's nominal power for discharging $P^{\textrm{EVC, nominal, discharging}}$  when $\text{if } a_{t - 1}^{\textrm{EVC}_{b\_n\_p}} < 0$ (\cref{eqn:charger_power_limits}). The supplied charging energy from the charger to the connected EV battery, for a controlled input power is then defined by \Cref{eqn:electric_vehicle_charger_output_values} as the product of the electricity consumption $E_t^{\textrm{EVC}_{b\_n\_p}}$ and technical efficiency $\eta^{\textrm{EVC, technical}}$.

\begin{multline}
    E_t^{\textrm{EVC}_{b\_n\_p}} = a_{t - 1}^{\textrm{EVC}_{b\_n\_p}} \\ \times P^{\textrm{EVC}_{b\_n\_p}, \textrm{nominal, charging | discharging}}
    \label{eqn:charger_power_limits}
\end{multline}

\begin{equation}
    Q_t^{\textrm{EVC}_{b\_n\_p}} = \eta^{\textrm{EVC}_{b\_n\_p}, \textrm{technical}} \times E_t^{\textrm{EVC}_{b\_n\_p}}
    \label{eqn:electric_vehicle_charger_output_values}
\end{equation}

\textbf{Electric vehicle:} \label{sec:environment-electric_vehicle_model}
The \gls{ev} model replicates the real-world operational attributes and practical constraints of \glspl{ev} within a simulation environment, specifically, those acting as \gls{der} within the system, e.g., participating in one of the defined modes. \Glspl{ev} in the simulation can be connected to an EVC (\Cref{sec:environment-electric_vehicle_charger}) and consume energy (in all modes) and discharge back to the grid (in V2G mode). \Glspl{ev} will connect and disconnect from the chargers as per the modes pre-simulated file described at \cref{sec:environment-electric_vehicle_dynamics}. The \gls{ev} uses the same battery model as the stationary \gls{bess} (\Cref{sec:environment-battery_model}).

\textbf{Modes:} \label{sec:environment-electric_vehicle_dynamics}
In the simulation modes for \gls{ev} energy management, we establish that the control over \gls{ev} charging is exercised exclusively when the vehicles are plugged into a building's EVC. Specifically, energy management algorithms, referred to as agents, regulate the power of the chargers, which in turn charges/discharges the connected \glspl{ev}. It is important to note that these agents will not have an influence on the routine when an \gls{ev} arrives or leaves; these events are dictated by the real-life schedules and habits of the vehicle owners, which are pre-simulated for CityLearnV2. To simulate this mode, the concept of energy flexibility is utilized, adapted from the standardized FlexOffer (FO) model in \cite{pedersenflexoffers}.

\subsubsection{Photovoltaic system} \label{sec:environment-pv}
The \gls{pv} model makes use of pre-simulated inverter output power per unit of installed \gls{pv} capacity, reported as a time series, $P_t^{\textrm{I}}$. For \gls{pv} generation estimates, the System Advisory Model (SAM) \cite{blair2014system} is used to generate location specific data. The \gls{pv} generation is thus the negative product of a user-defined nominal power, $P^{\textrm{PV, nominal}}$ and $P_t^{\textrm{I}}$ (\cref{eqn:photovoltaic_system-electricity_consumption}).

\begin{equation}
    E_t^{\textrm{PV}} = -P^{\textrm{PV, nominal}} \times P_t^{\textrm{I}}
    \label{eqn:photovoltaic_system-electricity_consumption}
\end{equation}
\subsection{Thermal dynamics model} \label{sec:environment-temperature_dynamics_model_methodology}
\subsubsection{Fitting data generation} \label{sec:environment-temperature_dynamics_model-fitting_data_generation}
To generate the \gls{lstm} fitting data, we first build an EnergyPlus model of the building or source a representative model from open-source datasets like \cite{wilson_end-use_2022}. The energy model is paired with a weather file for the desired building location and simulated multiple times to obtain building ideal cooling and heating loads as well as the effect supplying thermal energy below or above the ideal loads has on the indoor dry-bulb temperature. The first simulation (\textit{energyplus.simulation.1}) uses the as-provided mechanical \gls{hvac} system in the energy model as a reference point for subsequent simulations.

A second simulation (\textit{energyplus.simulation.2}) where the mechanical \gls{hvac} system is replaced with a \textit{HVACTemplate:Zone:IdealLoadsAirSystem} EnergyPlus object which is an artificial \gls{hvac} system that completely supplies cooling or heating air to a building in sufficient quantity and 100\% efficiency to meet the building load in full (ideal load). We run the ideal load simulation to determine the ideal cooling and heating loads that satisfy the indoor dry-bulb temperature setpoint, and the result from this simulation is validated against that of \textit{energyplus.simulation.1}. The validation ensures that the as-provided building energy model integrity is not tampered with when replacing mechanical systems with ideal load system. Our check for integrity is that for the ideal load case, cooling and heating energy delivered to the building is equal or slightly less than that for the mechanical case. Indoor dry-bulb temperature in the building using the ideal load system should be similar to that from using the mechanical system.

The remaining simulations make up the \gls{lstm} training dataset where we replace the ideal load system with an \textit{OtherEquipment} EnergyPlus object which is a generic internal heat gain and loss equipment to simulate under-cooling, under-heating, over-cooling and overheating with respect to the building's ideal loads. The first of these simulations (\textit{energyplus.simulation.3}) supplies the ideal load from \textit{energyplus.simulation.2} to validate the operation of the generic equipment compared to the ideal load system, as the effect on temperature should closely match. The next simulation (\textit{energyplus.simulation.4}) free-floats the temperature by supplying neither cooling nor heating energy to the building while the remaining simulations (\textit{energyplus.simulation.5-n}) have the ideal load varied using the inequality, $A \cdot I_t \le I_t \le B \cdot I_t$ with a probability, $p^{L_t \neq I_t}$, that the ideal load is not used. $A$ and $B$ are coefficients that define the variation range where $p^{L_t \neq I_t}=0.6$, $A=0.3$ and $B=1.7$ have been used in practice \cite{deltetto_data-driven_2020}.

Finally, we query the results of the generic internal heat gain/loss equipment simulations to extract features that make up the \gls{lstm} fitting data, including the following variables: direct and diffuse solar irradiance, outdoor dry-bulb temperature, occupant count, cooling and heating loads, month, day-of-week, hour, and indoor dry-bulb temperature (target variable). Note that while we use EnergyPlus for our simulations, these data can be generated with any other \gls{beps} software.

\subsubsection{Training and testing} \label{sec:environment-temperature_dynamics_model-training_and_testing}
We split the generated data into training, validation and testing sets. The method of splitting ensure that each of the simulations with reference \textit{energyplus.simulation.3-n} are equally represented in each set. The sliding window method is used in training where we partition the training data into sequences of $l$ time steps. $l=12$ has been used in practice \cite{pinto_data-driven_2021}. Each sequence represents a window of independent features. Each sequence is then paired with its corresponding target, forming input-output pairs. By sliding the window along the time axis, multiple input-output pairs are generated, covering the entirety of the training data. Through this iterative process, the \gls{lstm} model learns to predict the indoor dry-bulb temperature at a given time step, $t$, leveraging the lagged input sequence leading up to the previous $t - l$ time steps. This enables the model to effectively capture temporal dynamics based on past observations.

We test the model in a closed loop, whereby the model's predictions are used as input for subsequent predictions. The model's performance is evaluated on its minimization of its \gls{rmse} (\cref{eqn:root_mean_square_error}) and \gls{mape} (\cref{eqn:mean_absolute_percentage_error}) where $n$ is the number of time steps in the testing period, $T_t^{\textrm{in}}$ is the true temperature at time step $t$ and $\hat{T}_t^{\textrm{in}}$ is the predicted temperature.

\begin{equation}
    \textrm{RMSE} = \sqrt{\frac{\sum_{t=0}^{n - 1}{\Big(T_t^{\textrm{in}} - \hat{T}_t^{\textrm{in}}\Big)^2}}{n}}
    \label{eqn:root_mean_square_error}
\end{equation}

\begin{equation}
    \textrm{MAPE} = \frac{1}{n} \times \sum_{t=0}^{n - 1}{\Bigg \lvert \frac{T_t^{\textrm{in}} - \hat{T}_t^{\textrm{in}}}{T_t^{\textrm{in}}} \Bigg \rvert}
    \label{eqn:mean_absolute_percentage_error}
\end{equation}

\subsection{Observation space} \label{sec:control-observation_space}
There are over 50 observable continuous states in CityLearn as summarized in \cref{tab:citylearn_observation_space} and categorized as one of the following: calendar, weather, environmental, cost, load, consumption, flexibility, efficiency, comfort, occupancy, and resilience observations. The weather and cost observations include 6, 12, and 24 hour forecasts that can inform future self-generation capability, loads and electric device efficiency. In ideal load control, the load and consumption observations inform the building loads and the energy used to satisfy the loads whereas in a controlled load scenario, where the load observations are calculated, the load and consumption observations inform the impact of the agent(s) control action(s) on energy use. Given that simultaneous cooling and heating is not possible in a building, the \verb|hvac_mode| is used to determine which of these two thermal modes a building is operating in at any given time step. The thermal loads can also be unmet and the indoor dry-bulb temperature allowed to free-float when \verb|hvac_mode=off|. The flexibility observations provide the control agent with the building's capacity for upward or downward energy flexibility. There are as many \gls{ev}-related observations as there are \glspl{ev} chargers in a building. The efficiency observations are the \gls{cop} of the cooling, heating and \gls{dhw} devices when they are heat pump type otherwise, it is their technical efficiency e.g., electric heater. The comfort observations provide information on occupant thermal and humidity comfort as well as occupant thermostat interaction. \verb|occupant_count| specifies if the building is occupied or not and the number of people during occupancy. The \verb|power_outage| observation is the power outage signal discussed in \cref{sec:environment-power_outage_model}.

Observations are (1) either shared i.e., equal value in all buildings or building specific, (2) either control action dependent or control agnostic, and (3) either statically defined in a time series file or calculated during runtime. The inclusion of an observation in a control problem's observation space is dependent on the control problem formulation including the controlled \glspl{der} and control objective and is customizable by the user.

\begin{table*}[]
    \centering
    \footnotesize
    \caption{Observation space.}
    \label{tab:citylearn_observation_space}
    \begin{tabular}{llccl}
        \toprule
        \textbf{Name} & \textbf{Unit} & \textbf{Shared} & \textbf{Control-dependent} & \textbf{Source} \\
        \midrule
        \textbf{Calendar} \\
        \verb|month| & - & \ding{51} & & \verb|building_id.csv| \\
        \verb|day_type| & - & \ding{51} & & \verb|building_id.csv| \\
        \verb|hour| & - & \ding{51} & & \verb|building_id.csv| \\
        \verb|daylight_savings_status| & true/false & \ding{51} & & \verb|building_id.csv| \\
        \textbf{Weather} \\
        \verb|outdoor_dry_bulb_temperature|\textsubscript{$t + (0, 6, 12, 24)$} & $^{\circ}$C & \ding{51} & & \verb|weather.csv| \\
        \verb|outdoor_relative_humidity|\textsubscript{$t + (0, 6, 12, 24)$} & \% & \ding{51} & & \verb|weather.csv| \\
        \verb|diffuse_solar_irradiance|\textsubscript{$t + (0, 6, 12, 24)$} & W/m\textsuperscript{2} & \ding{51} & & \verb|weather.csv| \\
        \verb|direct_solar_irradiance|\textsubscript{$t + (0, 6, 12, 24)$} & W/m\textsuperscript{2} & \ding{51} & & \verb|weather.csv| \\
        \textbf{Environmental} \\
        \verb|carbon_intensity| & kgCO\textsubscript{2}e/kWh & \ding{51} & & \verb|carbon_intensity.csv| \\
        \textbf{Cost} \\
        \verb|electricity_pricing|\textsubscript{$t + (0, 6, 12, 24)$} & \$/kWh & & & \verb|pricing.csv| \\
        \textbf{Load} \\
        \verb|cooling_demand| & kWh & & \ding{51} & \verb|building_id.csv|/calculated \\
        \verb|heating_demand| & kWh & & \ding{51} & \verb|building_id.csv|/calculated \\
        \verb|dhw_demand| & kWh & & & \verb|building_id.csv| \\
        \textbf{Consumption} \\
        \verb|cooling_electricity_consumption| & kWh & & \ding{51} & calculated \\
        \verb|heating_electricity_consumption| & kWh & & \ding{51} & calculated \\
        \verb|dhw_electricity_consumption| & kWh & & \ding{51} & calculated \\
        \verb|electrical_storage_electricity_consumption| & kWh & & \ding{51} & calculated \\
        \verb|non_shiftable_load| & kWh & & & \verb|building_id.csv| \\
        \verb|net_electricity_consumption| & kWh & & \ding{51} & calculated \\
        \verb|hvac_mode| & cooling/heating/off & & & \verb|building_id.csv| \\
        \textbf{Flexibility} \\
        \verb|cooling_storage_soc| & - & & \ding{51} & calculated \\
        \verb|heating_storage_soc| & - & & \ding{51} & calculated \\
        \verb|dhw_storage_soc| & - & & \ding{51} & calculated \\
        \verb|electrical_storage_soc| & - & & \ding{51} & calculated \\
        \verb|electrical_vehicle_soc| & - & & \ding{51} & calculated \\
        \verb|electrical_vehicle_estimated_arrival_soc| & - & & & \verb|electric_vehicle_id.csv| \\
        \verb|electrical_vehicle_required_departure_soc| & - & & & \verb|electric_vehicle_id.csv| \\
        \verb|electrical_vehicle_estimated_arrival_time| & - & & & \verb|electric_vehicle_id.csv| \\
        \verb|electrical_vehicle_estimated_departure_time| & - & & & \verb|electric_vehicle_id.csv| \\
        \verb|electrical_vehicle_charger_state| & - & & & \verb|electric_vehicle_id.csv| \\
        \verb|solar_generation| & kWh & & & calculated \\
        \textbf{Efficiency} \\
        \verb|cooling_device_efficiency| & - & & & calculated \\
        \verb|heating_device_efficiency| & - & & & calculated \\
        \verb|dhw_device_efficiency| & - & & & calculated \\
        \textbf{Comfort} \\
        \verb|indoor_dry_bulb_temperature| & $^{\circ}$C & & \ding{51} & \verb|building_id.csv|/calculated \\
        \verb|indoor_dry_bulb_temperature_set_point| & $^{\circ}$C & & \ding{51} & \verb|building_id.csv|/calculated \\
        \verb|indoor_dry_bulb_temperature_delta| & $^{\circ}$C & & \ding{51} & calculated \\
        \verb|indoor_dry_bulb_temperature_set_point_override_delta| & $^{\circ}$C & & \ding{51} & calculated \\
        \verb|indoor_relative_humidity| & \% & & & \verb|building_id.csv| \\
        \textbf{Occupancy} \\
        \verb|occupant_count| & people & & & \verb|building_id.csv| \\
        \textbf{Resilience} \\
        \verb|power_outage| & true/false & & & \verb|building_id.csv|/calculated \\
        \bottomrule
    \end{tabular}
\end{table*}
\subsection{Action space} \label{sec:control-action_space}
\Cref{tab:citylearn_action_space} summarizes the continuous action space where there are five \gls{ess}-related actions controlling the proportion of storage capacity to be charged or discharged and two \gls{hvac} electric device actions controlling the proportion of nominal power to be supplied. There are as many \verb|electric_vehicle_storage| actions as there are \glspl{ev} chargers in a building. Depending on the control problem and the \glspl{der} in a building, some actions may be excluded in the action space.

\begin{table*}[]
    \centering
    \footnotesize
    \caption{Action space.}
    \label{tab:citylearn_action_space}
    \begin{tabularx}{\textwidth}{lrX}
        \toprule
        \textbf{Name} & \textbf{$a_t$ range} & \textbf{Description} \\
        \midrule
        \textbf{Energy storage system} \\
        \verb|cooling_storage| & [-1, 1] & Proportion of \verb|cooling_storage| capacity to be charged ($a_t > 0$) or discharged ($a_t < 0$). \\
        \verb|heating_storage| & [-1, 1] & Proportion of \verb|heating_storage| capacity to be charged ($a_t > 0$) or discharged ($a_t < 0$). \\
        \verb|dhw_storage| & [-1, 1] & Proportion of \verb|dhw_storage| capacity to be charged ($a_t > 0$) or discharged ($a_t < 0$). \\
        \verb|electrical_storage| & [-1, 1] & Proportion of \verb|electrical_storage| capacity to be charged ($a_t > 0$) or discharged ($a_t < 0$). \\
        \verb|electric_vehicle_storage| & [-1, 1] & Proportion of \verb|electric_vehicle_storage| capacity to be charged ($a_t > 0$) or discharged ($a_t < 0$). \\
        \textbf{Electric device} \\
        \verb|cooling_device| & [0, 1] & Proportion of \verb|cooling_device| nominal power to be supplied. \\
        \verb|heating_device| & [0, 1] & Proportion of \verb|heating_device| nominal power to be supplied. \\
        \bottomrule
    \end{tabularx}
\end{table*}
\subsection{Reward functions} \label{sec:control-reinforcement_learning_control_reward_functions}
\Cref{tab:citylearn_reward_functions} summarizes four reward functions defined in CityLearn for \gls{rlc} although, other user-defined reward functions are possible depending on the control problem formulation. \Cref{eqn:reward_function-electricity_consumption,eqn:reward_function-marl,eqn:reward_function-solar_penalty,eqn:reward_function-comfort} define these reward functions at the building-level, $r_t^{\textrm{building}}$, and for centralized control architecture, the district reward, $r_t^{\textrm{district}}$, is the sum of the building-level rewards (\cref{eqn:district_reward}).

The reward function guides the learning process in \gls{rlc} and is akin to the objective in \gls{mpc}.

\begin{equation}
    r_t^{\textrm{district}} = \sum_{i=0}^{b-1}{r_t^{\textrm{building } i}}
    \label{eqn:district_reward}
\end{equation}

\begin{table*}[]
    \caption{Reinforcement learning reward functions in CityLearn.}
    \label{tab:citylearn_reward_functions}
    \footnotesize
    \centering
    \begin{tblr}{lXXr}
        \hline
        \textbf{Name} & \textbf{Description} & \textbf{$r_t^{\textrm{building}}$} \\
        \hline
        Electricity consumption & Encourages reduced consumption from the grid and does not reward nor penalize net export i.e., ($E_t^{\textrm{building, net}} < 0$). Exponent, $a$, used to impose greater penalty for $E_t^{\textrm{building, net}} >> 0$. & $-\textrm{max}\left(E_t^{\textrm{building, net}}, 0\right)^a$ & \Eq\label{eqn:reward_function-electricity_consumption} \\
        MARL & Multi-agent reward used to share information between the agents and rewards them for reducing the district net electricity consumption, $E_t^{\textrm{district, net}}$ \cite{vazquez-canteli_marlisa_2020}. For a building, the value is (1) negative if the building is consuming electricity from the grid ($E_t^{\textrm{building, net}} > 0$) while the district is also consuming electricity from the grid ($E_t^{\textrm{district, net}} > 0$), (2) positive when the building generates more electricity than it consumes ($E_t^{\textrm{building, net}} < 0$) though the district is consuming electricity from the main grid ($E_t^{\textrm{district, net}} > 0$), since the building is contributing to making the district self-sufficient. If the district is self-sufficient i.e., $E_t^{\textrm{district, net}} < 0$, the reward is 0. & $0.01 \times \left({E_t^{\textrm{building, net}}}^2\right) \times \textrm{max}\left(E_t^{\textrm{district, net}}, 0\right) \times \begin{cases} 1 & \text{ if } E_t^{\textrm{building, net}} < 0 \\ -1 & \text{ otherwise} \end{cases}$ & \Eq\label{eqn:reward_function-marl} \\
        Solar penalty & Encourages net-zero energy use by penalizing load satisfaction from the grid when there is stored energy in \glspl{ess} as well as penalizing net export when \glspl{ess} are not fully charged \cite{bs2023_1404}. There is neither penalty nor reward when \glspl{ess} are fully charged during net export. Whereas, when \glspl{ess} are charged to capacity and there is net import the penalty is maximized. & $\sum_{i=0}^n - \Bigg( \Bigg(1 + \frac{E_t^{\textrm{building, net}}}{\left|E_t^{\textrm{building, net}}\right|} \times \textrm{ESS}_{t}^{\textrm{i, SoC}} \Bigg) \times \left|E_t^{\textrm{building, net}}\right| \Bigg)$ & \Eq\label{eqn:reward_function-solar_penalty} \\
        Comfort & Negative difference between indoor dry-bulb temperature, $T_t^{\textrm{in}}$, and setpoint, $T_t^{\textrm{spt}}$, raised to some exponent,$a | b$, ($a \le b$) if outside the comfort band, $T^{\Delta}$ \cite{pinto_data-driven_2021}. If within the comfort band, the reward is the negative difference when in cooling mode and temperature is below the setpoint or when in heating mode and temperature is above the setpoint. The reward is 0 if within the comfort band and above the setpoint in cooling mode or below the setpoint and in heating mode. The exponents teach the control agent to take actions that use the least energy while keeping $T_t^{\textrm{in}}$ close to $T_t^{\textrm{spt}}$. & $\begin{cases}
        -\left \lvert T_t^{\textrm{in}} - T_t^{\textrm{spt}} \right \rvert ^b & \text{ if } T_t^{\textrm{in}} < \left(T_t^{\textrm{spt}} - T^{\Delta}\right) \\ & \text{\& hvac\_mode = cooling} \\ -\left \lvert T_t^{\textrm{in}} - T_t^{\textrm{spt}} \right \rvert ^a & \text{ if } T_t^{\textrm{in}} < \left(T_t^{\textrm{spt}} - T^{\Delta}\right) \\ & \text{\& hvac\_mode = heating} \\ -\left \lvert T_t^{\textrm{in}} - T_t^{\textrm{spt}} \right \rvert & \text{ if }  \left(T_t^{\textrm{spt}} - T^{\Delta}\right) \le T_t^{\textrm{in}} < T_t^{\textrm{spt}} \\ & \text{\& hvac\_mode = cooling} \\ 0 & \text{ if }  \left(T_t^{\textrm{spt}} - T^{\Delta}\right) \le T_t^{\textrm{in}} < T_t^{\textrm{spt}} \\ & \text{\& hvac\_mode = heating} \\ 0 & \text{ if }  T_t^{\textrm{spt}} \le T_t^{\textrm{in}} \le \left(T_t^{\textrm{spt}} + T^{\Delta}\right) \\ & \text{\& hvac\_mode = cooling} \\ -\left \lvert T_t^{\textrm{in}} - T_t^{\textrm{spt}} \right \rvert & \text{ if }  T_t^{\textrm{spt}} \le T_t^{\textrm{in}} \le \left(T_t^{\textrm{spt}} + T^{\Delta}\right) \\ & \text{\& hvac\_mode = heating} \\ -\left \lvert T_t^{\textrm{in}} - T_t^{\textrm{spt}} \right \rvert ^a & \text{ if }  \left(T_t^{\textrm{spt}} + T^{\Delta}\right) < T_t^{\textrm{in}} \\ & \text{\& hvac\_mode = cooling} \\ -\left \lvert T_t^{\textrm{in}} - T_t^{\textrm{spt}} \right \rvert ^b & \text{ otherwise} \end{cases}$ & \Eq\label{eqn:reward_function-comfort} \\
        \hline
    \end{tblr}
\end{table*}
\subsection{Control algorithms} \label{sec:control-control_algorithms}
\Cref{tab:citylearn_control_agents} summarizes the control agent algorithms that are defined in CityLearn as well as supported standardized \gls{rlc} algorithm libraries. The Baseline agent is used to establish a \gls{bau} scenario without \glspl{der} control while the Random agent takes arbitrary actions. All \gls{rbc} agents in CityLearn are descendants of the Hour-\gls{rbc} agent that prescribes actions with respect to the hour-of-day. CityLearn provides three main types of model-free \gls{rlc} agents: Tabular Q-Learning \cite{richard_s_sutton_and_andrew_g_barto_reinforcement_2018} for discretized observation and action spaces, \gls{sac} \cite{haarnoja_soft_2018} and MARLISA \cite{vazquez-canteli_marlisa_2020} for continuous observation and action spaces. The \gls{sac} algorithm is adapted for either single-agent or independent multi-agent configurations while the MARLISA algorithm is for purely multi-agent configurations with the ability to share information amongst agents. \gls{sac}-\gls{rbc} and MARLISA-\gls{rbc} are sub-types of the \gls{sac} and MARLISA algorithms that make use of an Hour-\gls{rbc} for offline training to improve their learning \cite{nweye_real-world_2022}. For a wider range of algorithm options, we provide wrappers for interfacing with algorithms implemented in Stable-Baselines3 \cite{raffin_stable-baselines3_2021} and RLlib \cite{liang_rllib_2017} libraries.

Examples of how to use these control algorithms are provided in our web documentation \footnote{ANONYMOUS}

\begin{table*}[]
    \centering
    \caption{Control agents in CityLearn and supported third-party standardized control algorithm libraries.}
    \label{tab:citylearn_control_agents}
    \footnotesize
    \begin{tabularx}{\textwidth}{llcccX}
         \toprule
        \multirow{2}{*}{\textbf{Name}} & \multirow{2}{*}{\textbf{Theory}} & \multicolumn{3}{c}{\textbf{Configuration}} & \multirow{2}{*}{\textbf{Description}} \\
         & & \textbf{Single-agent} & \textbf{Independent} & \textbf{Coordinated} & \\
        \midrule
        \textbf{CityLearn} \\
        Baseline & - & \ding{51} & \ding{51} & & \Gls{bau} scenario where neither \glspl{ess} nor electric devices are controlled and ideal loads are satisfied. \\
        Random & - & \ding{51} & \ding{51} & & Arbitrarily selected actions. \\
        Hour-RBC & \gls{rbc} & \ding{51} & \ding{51} & & Prescribes control actions based on custom hour-of-use values. \\
        Tabular Q-Learning & \gls{rlc} & \ding{51} & \ding{51} & & Model-free \gls{rlc} algorithm for low-dimensional discrete observation and action spaces \cite{richard_s_sutton_and_andrew_g_barto_reinforcement_2018}. \\
        SAC & \gls{rlc} & \ding{51} & \ding{51} & & Implementation of the model-free \gls{sac} algorithm \cite{haarnoja_soft_2018}. \\
        MARLISA & \gls{rlc} & & \ding{51} & \ding{51} & \Gls{sac} extension with shared reward and internal building model to predict district electricity consumption for cooperative control \cite{vazquez-canteli_marlisa_2020}. \\
        SAC-RBC & \gls{rbc}, \gls{rlc} & \ding{51} & \ding{51} & & \gls{sac} paired with an Hour-RBC for offline training or initial exploration \cite{nweye_real-world_2022}. \\
        MARLISA-RBC & \gls{rbc}, \gls{rlc} & & \ding{51} & \ding{51} & MARLISA paired with an Hour-RBC for offline training or initial exploration. \\
        \textbf{Third-party} \\
        Stable-Baselines3 wrapper & \gls{rlc} & \ding{51} & & & Interface for Stable-Baselines3 single-agent algorithms \cite{raffin_stable-baselines3_2021}. \\
        RLlib wrapper & \gls{rlc} & \ding{51} & \ding{51} & \ding{51} & Interface for RLlib single-agent and multi-agent algorithms \cite{liang_rllib_2017}. \\
        \bottomrule
    \end{tabularx}
\end{table*}
\subsection{Key performance indicators}
\Cref{eqn:kpi-discomfort} is a thermal comfort \gls{kpi} that quantifies the proportion of total simulation time steps, $n$, where the difference between a building's indoor dry-bulb temperature, $T_t^{\textrm{in}}$, and set point, $T_t^{\textrm{spt}}$, falls outside a comfort band, $T^{\Delta}$. \Cref{,eqn:kpi-cold_discomfort,eqn:kpi-hot_discomfort} are similar to \cref{eqn:kpi-discomfort} but consider when the difference is either below or above $T^{\Delta}$. \Cref{eqn:kpi-minimum_cold_delta,eqn:kpi-maximum_cold_delta,eqn:kpi-average_cold_delta,eqn:kpi-minimum_hot_delta,eqn:kpi-maximum_hot_delta,eqn:kpi-average_hot_delta} report the descriptive statistics of $T_t^{\textrm{in}} - T_t^{\textrm{spt}}$ for either cold discomfort ($T_t^{\textrm{in}} < T_t^{\textrm{spt}}$) or hot discomfort ($T_t^{\textrm{in}} > T_t^{\textrm{spt}}$) at the building level.

The energy \glspl{kpi} are defined in \cref{eqn:kpi-total_electricity_consumption,eqn:kpi-zero_net_energy,eqn:kpi-average_daily_peak,eqn:kpi-all_time_peak,eqn:kpi-average_ramping,eqn:kpi-one_minus_load_factor} where \cref{eqn:kpi-total_electricity_consumption} defines a building's total electricity consumption, $\textrm{max}\left(E_t^{\textrm{building, net}}, 0\right)$, from the grid (net import) whereas \cref{eqn:kpi-zero_net_energy} is a building's total net electricity consumption, $E_t^{\textrm{building, net}}$, (sum of import and export). \Cref{eqn:kpi-average_daily_peak} is a district leve \gls{kpi} quantifying an entire district's average daily peak net electricity consumption, $E^{\textrm{district, net}}$, where $h$ is the hours per day, and $d$ is the number of days. The peak district net electricity consumption at any time step is quantified by \cref{eqn:kpi-all_time_peak}. Ramping is defined in \cref{eqn:kpi-average_ramping} as the positive difference in $E^{\textrm{district, net}}$ between two consecutive time steps. It a district level \gls{kpi} that quantifies the smoothness of the district electricity consumption during the up-ramp that typically happens in the early evening when renewable power generation depletes and fossil-fueled power plants come online. \Cref{eqn:kpi-one_minus_load_factor} is also a district level \gls{kpi} that evaluates the average ratio of daily average and peak net electricity consumption, termed, load factor. The load factor is the efficiency of electricity consumption and is bound between 0 (highly inefficient) and 1 (highly efficient) thus, the goal is to maximize the load factor or minimize $1 - \textrm{load factor}$. These district-level energy \glspl{kpi} are evaluated at the building level by substituting $E_t^{\textrm{district, net}}$ with $E_t^{\textrm{building, net}}$ in their equations. 

\Cref{eqn:kpi-total_cost} and \cref{eqn:kpi-total_co2e_emissions} evaluate the cost and CO\textsubscript{2}e emissions from net import at the building level using time-dependent electricity rate, $R_t$, and emissions rate, $G_t$. 

\Cref{eqn:kpi-one_minus_thermal_resilience,eqn:kpi-total_unserved_energy} evaluate the resilience provided by a control algorithm during power outages i.e., $O_t > 0$, from thermal comfort and energy perspectives. Thermal resilience (\cref{eqn:kpi-one_minus_thermal_resilience}) is similar to \cref{eqn:kpi-discomfort} but is evaluated only during power outage events. Total unserved energy (\cref{eqn:kpi-total_unserved_energy}) quantifies the difference between the energy needed to satisfy building loads and the actual energy supplied to the building by the electric devices and \glspl{ess}.

\begin{table*}[]
    \caption{Summary of post-simulation \glspl{kpi} in CityLearn.}
    \label{tab:kpi}
    \footnotesize
    \centering
    \begin{tblr}{lccXr}
        \hline
        \SetCell[r=2,c=1]{l} \textbf{Name} & \SetCell[r=1,c=2]{c} \textbf{Spatial resolution} & & & \SetCell[r=2,c=1]{r} \textbf{Equation} \\
         & \textbf{Building} & \textbf{District} & & \\
        \hline
        \textbf{Thermal comfort} & & & & \\
        Discomfort time steps & \ding{51} & & $\frac{1}{n} \times \sum_{t=0}^{n-1}{\begin{cases} 1 & \text{ if } \left \lvert T_t^{\textrm{in}} - T_t^{\textrm{spt}} \right \rvert > T^{\Delta} \\ 0 & \text{ otherwise} \end{cases}}$ & \Eq\label{eqn:kpi-discomfort} \\
        Cold discomfort time steps & \ding{51} & & $\frac{1}{n} \times \sum_{t=0}^{n-1}{\begin{cases} 1 & \text{ if } T_t^{\textrm{in}} - T_t^{\textrm{spt}} < -T^{\Delta} \\ 0 & \text{ otherwise} \end{cases}}$ & \Eq\label{eqn:kpi-cold_discomfort} \\
        Hot discomfort time steps & \ding{51} & & $\frac{1}{n} \times \sum_{t=0}^{n-1}{\begin{cases} 1 & \text{ if } T_t^{\textrm{in}} - T_t^{\textrm{spt}} > T^{\Delta} \\ 0 & \text{ otherwise} \end{cases}}$ & \Eq\label{eqn:kpi-hot_discomfort} \\
        Minimum cold discomfort & \ding{51} & & $\textrm{min}\left(\left\lvert \textrm{min}\left(0, T_0^{\textrm{in}} - T_0^{\textrm{spt}}\right) \right\rvert, \dots,  \left\lvert \textrm{min}\left(0, T_{n-1}^{\textrm{in}} - T_{n-1}^{\textrm{spt}}\right) \right\rvert \right)$ & \Eq\label{eqn:kpi-minimum_cold_delta} \\
        Maximum cold discomfort & \ding{51} & & $\textrm{max}\Big(\left\lvert \textrm{min}\left(0, T_0^{\textrm{in}} - T_0^{\textrm{spt}}\right) \right\rvert, \dots,  \left\lvert \textrm{min}\left(0, T_{n-1}^{\textrm{in}} - T_{n-1}^{\textrm{spt}}\right) \right\rvert \Big)$ & \Eq\label{eqn:kpi-maximum_cold_delta} \\
        Average cold discomfort & \ding{51} & & $\frac{1}{n} \times \sum_{t=0}^{n-1}{\left\lvert \textrm{min}\left(0, T_t^{\textrm{in}} - T_t^{\textrm{spt}}\right) \right\rvert}$ & \Eq\label{eqn:kpi-average_cold_delta} \\
        Minimum hot discomfort & \ding{51} & & $\textrm{min}\Big(\textrm{max}\left(0, T_0^{\textrm{in}} - T_0^{\textrm{spt}} \right), \dots, \textrm{max}\left(0, T_{n-1}^{\textrm{in}} - T_{n-1}^{\textrm{spt}}\right) \Big)$ & \Eq\label{eqn:kpi-minimum_hot_delta} \\
        Maximum hot discomfort & \ding{51} & & $\textrm{max}\Big(\textrm{max}\left(0, T_0^{\textrm{in}} - T_0^{\textrm{spt}}\right), \dots, \textrm{max}\left(0, T_{n-1}^{\textrm{in}} - T_{n-1}^{\textrm{spt}}\right) \Big)$ & \Eq\label{eqn:kpi-maximum_hot_delta} \\
        Average hot discomfort & \ding{51} & & $\frac{1}{n} \times \sum_{t=0}^{n-1}{\textrm{max}\left(0, T_t^{\textrm{in}} - T_t^{\textrm{spt}}\right)}$ & \Eq\label{eqn:kpi-average_hot_delta} \\
        \textbf{Energy} & & & & \\
        Total electricity consumption & \ding{51} & & $\sum_{t=0}^{n-1}\textrm{max}\left(E_t^{\textrm{building, net}}, 0\right)$ & \Eq\label{eqn:kpi-total_electricity_consumption} \\
        Total net electricity consumption & \ding{51} & & $\sum_{t=0}^{n-1}E_t^{\textrm{building, net}}$ & \Eq\label{eqn:kpi-zero_net_energy} \\
        Average daily peak & & \ding{51} & $\frac{h}{n} \times \sum_{d=0}^{n \div h - 1} \textrm{max} \left (E_{d \cdot h}^{\textrm{district, net}}, \dots, E_{d \cdot h + h - 1}^{\textrm{district, net}} \right )$ & \Eq\label{eqn:kpi-average_daily_peak} \\
        All-time peak & & \ding{51} & $\textrm{max} \left (E_{0}^{\textrm{district, net}}, \dots, E_{n}^{\textrm{district, net}} \right )$ & \Eq\label{eqn:kpi-all_time_peak} \\
        Total ramping & & \ding{51} & $\sum_{t=1}^{n-1} \textrm{max}\left(0, E_{t}^{\textrm{district, net}} - E_{t - 1}^{\textrm{district, net}}\right)$ & \Eq\label{eqn:kpi-average_ramping} \\
        Average (1 - load factor) & & \ding{51} & $\frac{h}{n} \times \sum_{d=0}^{n \div h} 1 - \frac{ \left ( \sum_{t=d \cdot h}^{d \cdot h +  h - 1} E_{t}^{\textrm{district, net}} \right ) \div h }{ \textrm{max} \left (E_{d \cdot h}^{\textrm{district, net}}, \dots, E_{d \cdot h +  h - 1}^{\textrm{district, net}} \right ) }$ & \Eq\label{eqn:kpi-one_minus_load_factor} \\
        \textbf{Cost} & & & & \\
        Total energy cost & \ding{51} & & $\sum_{t=0}^{n-1}\textrm{max}\left(E_t^{\textrm{building, net}}, 0\right) \times R_t$ & \Eq\label{eqn:kpi-total_cost} \\
        \textbf{Environmental} & & & & \\
        Total CO\textsubscript{2}e emissions & \ding{51} & & $\sum_{t=0}^{n-1}\textrm{max}\left(E_t^{\textrm{building, net}}, 0\right) \times G_t$ & \Eq\label{eqn:kpi-total_co2e_emissions} \\
        \textbf{Resiliency and thermal comfort} & & & & \\
        1 - thermal resilience & \ding{51} & & $\frac{1}{n^{\textrm{power outage}}} \times \sum_{t=0}^{n-1}{\begin{cases} 1 & \text{ if } \left \lvert T_t^{\textrm{in}} - T_t^{\textrm{spt}} \right \rvert > T^{\Delta} \text{ \& } O_t > 0 \\ 0 & \text{ otherwise} \end{cases}}$ & \Eq\label{eqn:kpi-one_minus_thermal_resilience} \\
        \textbf{Resiliency and energy} & & & & \\
        Total unserved energy & \ding{51} & & $\begin{cases}\Big(Q_t^{\textrm{cooling\_demand}} + Q_t^{\textrm{heating\_demand}} + Q_t^{\textrm{dhw\_demand}} + Q_t^{\textrm{non\_shiftable\_load}}\\+ Q_t^{\textrm{electric\_vehicle\_charger}}\Big)^{\textrm{expected}}- \Big(Q_t^{\textrm{cooling\_device}} + Q_t^{\textrm{cooling\_storage}}\\+ Q_t^{\textrm{heating\_device}}\\+ Q_t^{\textrm{heating\_storage}} + Q_t^{\textrm{dhw\_device}}\\+ Q_t^{\textrm{dhw\_storage}} + Q_t^{\textrm{non\_shiftable\_load}} + Q_t^{\textrm{electric\_vehicle\_charger}}\Big)^{\textrm{actual}} & \text{ if } O_t > 0 \\ 0 & \text{ otherwise}\end{cases}$ & \Eq\label{eqn:kpi-total_unserved_energy} \\
        \hline
    \end{tblr}
\end{table*}
\begin{sidewaystable*}
    \begin{threeparttable}
    \centering
    \footnotesize
    \caption{CityLearn datasets.}
    \label{tab:citylearn_datasets}
    \begin{tabularx}{\textwidth}{llllrlccccccccccccccc}
        \toprule
        \multirow{2}{*}{\textbf{Name}\tnote{1,2}} & \multirow{2}{*}{\textbf{Loc.}} & \textbf{CZ}\tnote{3} & \multirow{2}{*}{\textbf{Date}\tnote{4}} & \multirow{2}{*}{\textbf{Bldgs.}} & \multirow{2}{*}{\textbf{Type}\tnote{5}} & \multicolumn{3}{c}{\textbf{Model}} & \multicolumn{2}{c}{\textbf{Signal}} & \multicolumn{4}{c}{\textbf{Load}} & \multicolumn{5}{c}{\textbf{ESS}} & \textbf{PV} \\
        & & & & & & \textbf{\rotatebox[origin=c]{90}{Occupant}} & \textbf{\rotatebox[origin=c]{90}{Thermal}} & \textbf{\rotatebox[origin=c]{90}{Outage}} & \textbf{\rotatebox[origin=c]{90}{Pricing}} & \textbf{\rotatebox[origin=c]{90}{Emissions}} & \textbf{\rotatebox[origin=c]{90}{Cooling}} & \textbf{\rotatebox[origin=c]{90}{Heating}} & \textbf{\rotatebox[origin=c]{90}{DHW}} & \textbf{\rotatebox[origin=c]{90}{Non-shift}} & \textbf{\rotatebox[origin=c]{90}{Cooling}} & \textbf{\rotatebox[origin=c]{90}{Heating}} & \textbf{\rotatebox[origin=c]{90}{DHW}} & \textbf{\rotatebox[origin=c]{90}{Electrical}} & \textbf{EV} & \\
        \midrule
        *\_2020\_climate\_zone\_1 & New Orleans, LA, US & 2A & Jan-Dec, 05 & 9 & Com & & & & & \ding{51} & & & \ding{51} & \ding{51} & \ding{51} & & \ding{51} & \ding{51} & & \ding{51} \\
        *\_2020\_climate\_zone\_2 & Atlanta, GA, US & 3A & Jan-Dec, 05 & 9 & Com & & & & & \ding{51} & & & \ding{51} & \ding{51} & \ding{51} & & \ding{51} & \ding{51} & & \ding{51} \\
        *\_2020\_climate\_zone\_3 & Nashville, TN, US & 4A & Jan-Dec, 05 & 9 & Com & & & & & \ding{51} & & & \ding{51} & \ding{51} & \ding{51} & & \ding{51} & \ding{51} & & \ding{51} \\
        *\_2020\_climate\_zone\_4 & Chicago, IL, US & 5A & Jan-Dec, 05 & 9 & Com & & & & & \ding{51} & & & \ding{51} & \ding{51} & \ding{51} & & \ding{51} & \ding{51} & & \ding{51} \\
        *\_2021 & New Orleans, LA, US & 2A & Jan-Dec 05, 14-17 & 9 & Com & & & & & \ding{51} & & & \ding{51} & \ding{51} & \ding{51} & & \ding{51} & \ding{51} & & \ding{51} \\
        *\_2022\_phase\_1 & Fontana, CA, US & 3B & Aug, 16-Jul, 17 & 5 & Res & & & & \ding{51} & \ding{51} & & & & \ding{51} & & & & \ding{51} & & \ding{51} \\
        *\_2022\_phase\_2 & Fontana, CA, US & 3B & Aug, 16-Jul, 17 & 5 & Res & & & & \ding{51} & \ding{51} & & & & \ding{51} & & & & \ding{51} & & \ding{51} \\
        *\_2022\_phase\_3 & Fontana, CA, US & 3B & Aug, 16-Jul, 17 & 7 & Res & & & & \ding{51} & \ding{51} & & & & \ding{51} & & & & \ding{51} & & \ding{51} \\
        *\_2022\_phase\_all & Fontana, CA, US & 3B & Aug, 16-Jul, 17 & 17 & Res & & & & \ding{51} & \ding{51} & & & & \ding{51} & & & & \ding{51} & & \ding{51} \\
        baeda\_3dem & Albuquerque, NM, US & 4B & Jun-Aug, TMY3 & 4 & Com & & \ding{51} & & \ding{51} & & \ding{51} & & \ding{51} & \ding{51} & \ding{51} & & \ding{51} & & & \ding{51} \\
        *\_2023\_phase\_1 & Austin, TX, US & 2A & Jun, 18 & 3 & Res & & \ding{51} & \ding{51} & \ding{51} & \ding{51} & \ding{51} & & \ding{51} & \ding{51} & & & \ding{51} & \ding{51} & & \ding{51} \\
        *\_2023\_phase\_2\_local\_evaluation & Austin, TX, US & 2A & Jun, 18 & 3 & Res & & \ding{51} & \ding{51} & \ding{51} & \ding{51} & \ding{51} & & \ding{51} & \ding{51} & & & \ding{51} & \ding{51} & & \ding{51} \\
        *\_2023\_phase\_2\_online\_evaluation\_1 & Austin, TX, US & 2A & Jun, 21-Aug, 21 & 3 & Res & & \ding{51} & \ding{51} & \ding{51} & \ding{51} & \ding{51} & & \ding{51} & \ding{51} & & & \ding{51} & \ding{51} & & \ding{51} \\
        *\_2023\_phase\_2\_online\_evaluation\_2 & Austin, TX, US & 2A & Jun, 21-Aug, 21 & 3 & Res & & \ding{51} & \ding{51} & \ding{51} & \ding{51} & \ding{51} & & \ding{51} & \ding{51} & & & \ding{51} & \ding{51} & & \ding{51} \\
        *\_2023\_phase\_2\_online\_evaluation\_3 & Austin, TX, US & 2A & Jun, 21-Aug, 21 & 3 & Res & & \ding{51} & \ding{51} & \ding{51} & \ding{51} & \ding{51} & & \ding{51} & \ding{51} & & & \ding{51} & \ding{51} & & \ding{51} \\
        *\_2023\_phase\_3\_1 & Austin, TX, US & 2A & Jun, 21-Aug, 21 & 6 & Res & & \ding{51} & \ding{51} & \ding{51} & \ding{51} & \ding{51} & & \ding{51} & \ding{51} & & & \ding{51} & \ding{51} & & \ding{51} \\
        *\_2023\_phase\_3\_2 & Austin, TX, US & 2A & Jun, 21-Aug, 21 & 6 & Res & & \ding{51} & \ding{51} & \ding{51} & \ding{51} & \ding{51} & & \ding{51} & \ding{51} & & & \ding{51} & \ding{51} & & \ding{51} \\
        *\_2023\_phase\_3\_3 & Austin, TX, US & 2A & Jun, 21-Aug, 21 & 6 & Res & & \ding{51} & \ding{51} & \ding{51} & \ding{51} & \ding{51} & & \ding{51} & \ding{51} & & & \ding{51} & \ding{51} & & \ding{51} \\
        tx\_travis\_county\_neighborhood & Travis Co., TX, US & 2A & Jun-Aug 18 & 100 & Res & & \ding{51} & & & & \ding{51} & & \ding{51} & \ding{51} & & & \ding{51} & \ding{51} & & \ding{51} \\
        ca\_alameda\_county\_neighborhood & Alameda Co., CA, US & 3C & Jun-Aug 18 & 73 & Res & & \ding{51} & & & & \ding{51} & & \ding{51} & \ding{51} & & & \ding{51} & \ding{51} & & \ding{51} \\
        vt\_chittenden\_county\_neighborhood & Chittenden Co., VT, US & 6A & Jan-Mar 18 & 43 & Res & & \ding{51} & & & & & \ding{51} & \ding{51} & \ding{51} & & & \ding{51} & \ding{51} & & \ding{51} \\
        quebec\_neighborhood & Quebec, Canada & 6A & Jan-Mar 20-22 & 10 & Res & \ding{51} & \ding{51} & & \ding{51} & & & \ding{51} & \ding{51} & \ding{51} & & & & & & \\
        ev\_dataset & Porto, Portugal & 3B & Jan-Dec 18-22 & 9 & Com, Res & & & & & & & & & \ding{51} & & & & & \ding{51} & \\
        \bottomrule
    \end{tabularx}
    \begin{tablenotes}
        \item [1] The datasets are available at ANONYMOUS.
       \item [2] * in the name is a placeholder for citylearn\_challenge.
       \item [3] Climate zone temperature-humidity types: 2A (hot-humid), 3A (warm-humid), 3B (warm-dry), 3C (warm-marine), 4A (mixed-humid), 4B (mixed-dry), 5A (cool-humid), and 6A (cold-humid).
       \item [4] The year is an abbreviation of 20XX.
       \item [5] Building archetypes: Commercial (Com) and Residential (Res).
     \end{tablenotes}
     \end{threeparttable}
\end{sidewaystable*}


\end{document}